\documentclass{article}

\usepackage[dandb,final]{neurips_2025}

\usepackage[utf8]{inputenc}
\usepackage[T1]{fontenc}
\usepackage{hyperref}
\usepackage{url}
\usepackage{booktabs}
\usepackage{amsfonts}
\usepackage{nicefrac}
\usepackage{microtype}
\usepackage{xcolor}

\usepackage{longtable}
\usepackage{rotating}
\usepackage[table]{xcolor}
\usepackage[most]{tcolorbox}
\usepackage{listings}
\usepackage{subfigure}
\usepackage{wrapfig}

\usepackage{enumitem}%
\usepackage[capitalize,noabbrev]{cleveref}

\lstdefinelanguage{yaml}{
    keywords={true,false,null,y,n,yes,no},
    keywordstyle=\color{blue}\bfseries,
    sensitive=true,
    comment=[l]{\#},
    commentstyle=\color{gray}\ttfamily,
    stringstyle=\color{red}\ttfamily,
    morestring=[b]',
    morestring=[b]",
    basicstyle=\ttfamily,
    showstringspaces=false,
    columns=fullflexible,
}

\definecolor{Baseline_RAW}{HTML}{73d2de}
\definecolor{Best_RAW}{HTML}{d81159}

\definecolor{color_blind_red}{HTML}{f2b3c3}
\definecolor{color_blind_green}{HTML}{66c2b3}

\colorlet{Baseline}{Baseline_RAW!15}
\colorlet{Best}{Best_RAW!10}
\colorlet{AverageRank}{gray!20}

\title{Enhancing Multilingual LLM Pretraining with Model-Based Data Selection}

\author{%
  Bettina Messmer\thanks{Equal contribution}\\
  EPFL\\
  \texttt{bettina.messmer@epfl.ch}
  \And
  Vinko Sabolčec$^*$\\
  EPFL\\
  \texttt{vinko.sabolcec@epfl.ch}
  \And
  Martin Jaggi\\
  EPFL\\
  \texttt{martin.jaggi@epfl.ch}
}

\begin{document}

\maketitle

\begin{abstract}
Dataset curation has become a basis for strong large language model (LLM) performance.
While various rule-based filtering heuristics exist for English and multilingual datasets, model-based filtering techniques have primarily focused on English.
To address the disparity stemming from limited research on non-English languages, we develop a model-based filtering framework for multilingual datasets that aims to identify a diverse set of structured and knowledge-rich samples.
Our approach emphasizes transparency, simplicity, and efficiency, leveraging Transformer- and FastText-based classifiers to ensure the broad accessibility of our technique and data.
We conduct comprehensive ablation studies on the FineWeb-2 web crawl dataset across diverse language families, scripts, and resource availability to demonstrate the effectiveness of our method.
Training a 1B-parameter Llama model for 70B and 119B tokens, our approach can match the baseline MMLU score with as little as 15\% of the training tokens, while also improving across other benchmarks and mitigating the curse of multilinguality.
These findings provide strong evidence for the generalizability of our approach to other languages. As a result, we extend our framework to 20 languages for which we release the refined pretraining datasets.
\end{abstract}

\section{Introduction}
Large Language Models (LLMs) have demonstrated impressive performance improvements when trained on increasingly larger datasets and model sizes~\citep{brown2020language}. While~\citet{brown2020language} already observed the importance of using a cleaned version of Common Crawl for improved performance, the high cost of LLM training has further motivated research into better pretraining quality filters.

Deduplication and heuristic-based dataset cleaning have become standard practices in data curation~\citep{rae2021scaling, raffel2020exploring, de2024new}.
These quality filters are often complemented by additional filters, such as the removal of personally identifiable information (PII)~\citep{penedo2024fineweb} or model-based toxicity filtering~\citep{soldaini2024dolma}.
Recently, model-based filtering has also emerged as a promising method for quality filtering.
The release of FineWeb-Edu~\citep{penedo2024fineweb} demonstrated that pretraining on just 10\% of the tokens (38B) from an English dataset filtered using a model-based approach can achieve performance comparable to models trained on 350B tokens of unfiltered data.
Moreover, when trained on equivalent amounts of data, this model largely outperforms the baseline. Concurrently, the release of DataComp-LM (DCLM)~\citep{li2024datacomp} showed that competitive performance can be achieved using a simple and efficient model-based approach, namely a FastText~\citep{joulin2017bag} classifier trained on a carefully selected training dataset.

However, these recent advances have primarily focused on English data. This emphasis risks further widening the disparity in LLM performance between  languages, as less than half of internet content is written in English\footnote{\href{https://w3techs.com/technologies/overview/content_language}{w3techs.com/technologies/overview/content\_language}}. To address this concern, we aim to extend model-based filtering frameworks to multilingual datasets. While model perplexity-based filtering is commonly applied to multilingual datasets~\citep{wenzek2019ccnet, laurenccon2022bigscience, nguyen2023culturax}, the current state-of-the-art, FineWeb-2~\citep{penedo2024fineweb-2}, primarily relies on heuristic-based filters. In this work, we focus on model-based filtering with a quality definition that emphasizes: 1) structured data and 2) knowledge-rich data samples, to enhance multilingual pretraining datasets.

    \begin{wrapfigure}{r}{0.5\linewidth}
        \vspace{-1em}
        \centering
        \includegraphics[width=0.95\linewidth]{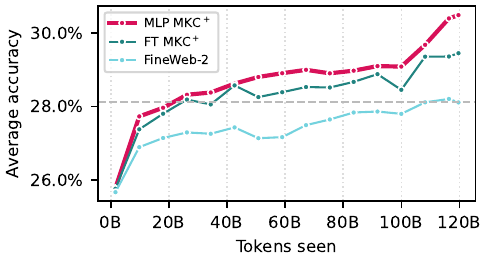}
        \vspace{-1em}
        \caption{Average accuracy on Chinese (CMMLU), German (MMLU), and French (MMLU) benchmarks during training: FineWeb-2 baseline compared to our methods (10\% data retention).}
        \label{fig:mmlu_avg_cmn_deu_fra}
        \vspace{-2mm}
    \end{wrapfigure}

To achieve this, we leverage embedding-based classification models. Firstly, we adopt the FastText quality filtering approach from DCLM to develop a unified framework for multilingual datasets that span diverse language families, scripts, and resource availability, focusing on Chinese, German, French, Arabic, and Danish as representative languages for our experiments.
Additionally, we extend this embedding-based approach by incorporating Transformer~\citep{vaswani2023attentionneed} embeddings, specifically XLM-RoBERTa~\citep{conneau2020unsupervisedcrosslingualrepresentationlearning}, for filtering. Figure~\ref{fig:mmlu_avg_cmn_deu_fra} shows the clear performance gains of our best FastText and Transformer embedding-based approaches over the state-of-the-art baseline FineWeb-2 data.

In summary, our contributions are as follows:
\begin{itemize}[itemsep=1pt,topsep=0pt]
    \item We develop a transparent, simple, and unified framework for multilingual model-based filtering at web scale, enabling data curation across diverse language families, scripts and resource availability.
    \item We present comprehensive per-language ablation studies of embedding-based multilingual quality filtering on top of the FineWeb-2 dataset~\citep{penedo2024fineweb-2}, achieving performance comparable to the baseline while using as little as 15\% of the tokens. We additionally analyze the impact of dataset contamination. Lastly, our experiments show that our dataset doesn't suffer from the \emph{curse of multilinguality}~\citep{chang2023multilingualitycurselanguagemodeling}.
    \item We evaluate the impact of different data selection classifiers, in particular  their training datasets, on the downstream performance of LLMs.
    \item We release the refined pretraining dataset\footnote{\href{https://huggingface.co/datasets/epfml/FineWeb2-HQ}{huggingface.co/datasets/epfml/FineWeb2-HQ}} covering 20 languages\footnote{Russian, Chinese, German, Japanese, Spanish, French, Italian, Portuguese, Polish, Dutch, Indonesian, Turkish, Czech, Vietnamese, Swedish, Persian, Arabic, Greek, Danish, Hungarian (dataset details in Appendix~\ref{app:sec:dataset})}, filtered using our proposed framework, along with the codebase\footnote{\href{https://github.com/epfml/fineweb2-hq}{{github.com/epfml/fineweb2-hq}}}, to advance multilingual language modeling.
\end{itemize}

\section{Related Work}
\textbf{Data Curation.} In order to pretrain LLMs on a large amount of diverse texts, Common Crawl\footnote{\href{https://commoncrawl.org/}{commoncrawl.org}} is often used as the base dataset. However, early works already observed that performing data curation on Common Crawl is crucial for model performance~\citep{brown2020language}.
In fact, data curation is important across NLP tasks~\citep{peter2023there,finkelstein2024introducing}. Specifically for pretraining data, there exist various data curation approaches, such as deduplication~\citep{lee-etal-2022-deduplicating}, PII removal~\citep{subramani-etal-2023-detecting}, or toxicity filtering~\citep{arnett2024toxicity}. Another important aspect is quality filtering of the documents. For this, the definition of quality is an important aspect. A common approach is to use heuristics to remove documents outside of the target distribution, such as filtering based on average word length, existence of punctuation, or document length~\citep{rae2021scaling, raffel2020exploring}. Another approach is to define model-based filters, where research has focused on perplexity measure of the text~\citep{wenzek2019ccnet,marion2023moreinvestigatingdatapruning,ankner2024perplexedperplexityperplexitybaseddata}, distributional similarity measures~\citep{li2024datacomp} and LLM-based quality assessment~\citep{gunasekar2023textbooksneed,wettig2024quratingselectinghighqualitydata,sachdeva2024traindataefficientllms,penedo2024fineweb}.
In this work, we build upon previous curated datasets based on heuristic filtering, namely the state-of-the-art dataset FineWeb-2~\citep{penedo2024fineweb-2}, and focus on model-based filtering for structured and knowledge-rich documents relying on textual embeddings.

\textbf{Curated English datasets.} One of the early curated datasets was C4~\citep{raffel2020exploring}, followed by MassiveText~\citep{rae2021scaling}. RefinedWeb~\citep{penedo2023refinedweb} was an important step forward, demonstrating that filtered web data can outperform selected high-quality data sources. Although these datasets have not been made fully publicly available, their filtering techniques have been expanded upon in recent fully public datasets, such as Dolma~\citep{soldaini2024dolma}, FineWeb, FineWeb-Edu~\citep{penedo2024fineweb} and DCLM~\citep{li2024datacomp}. While FineWeb primarily relies on filter heuristics for data quality, Dolma adopts model perplexity filtering. FineWeb-Edu takes model-based filtering a step further and relies on LLM-based quality assessment. DCLM, a concurrent work, has achieved competitive performance using a FastText~\citep{joulin2017bag} classifier trained on a carefully selected training dataset. In this work we adapt and extend this approach to the multilingual context.

\textbf{Curated Multilingual Datasets.}
Analogously to English datasets, significant work has been done in the multilingual space. For example, CCNet~\citep{wenzek2019ccnet} has been influential, with its language identification and model perplexity filtering for data quality being adopted in subsequent datasets. Similar to earlier English datasets, CCNet was not published directly, but rather provided tools for data cleaning. RedPajama~\citep{together2023redpajama} is a prominent multilingual dataset relying on these filtering techniques, offering data in 5 European languages. Other datasets, such as OSCAR~\citep{OrtizSuarezSagotRomary2019,AbadjiOrtizSuarezRomaryetal.2021,2022arXiv220106642A}, mC4~\citep{xue-etal-2021-mt5}, ROOTS~\citep{laurenccon2022bigscience}, MADLAD-400~\citep{kudugunta2023madlad400multilingualdocumentlevellarge}, CulturaX~\citep{nguyen2023culturax}, and HPLT~\citep{de-gibert-etal-2024-new-massive}, expanded coverage across a variety of language families and scripts. These datasets offer refined content for hundreds of languages, while FineWeb-2~\citep{penedo2024fineweb-2} pushes the limit to thousands of languages and further improves performance.
A concurrent work by \citet{martins2025eurollm} uses translation to train a multilingual quality filter based on the English FineWeb-Edu~\citep{penedo2024fineweb} scores. Our work similarly focuses on filtering high-quality samples across various language families and scripts.
However, we take a different approach: we train a separate classifier for each language from scratch using structured and knowledge-rich representative samples.
We limit our scope to 20 languages as the number of documents drops quickly and there is trade-off between retaining a sufficient number of pretraining tokens and ensuring data quality~\citep{muennighoff2023scalingdataconstrainedlanguagemodels, held2025optimizing}.

\textbf{Multilingual Embedding Models.} Early word embedding models like Word2Vec~\citep{mikolov2013efficientestimationwordrepresentations} and GloVe~\citep{pennington-etal-2014-glove} lacked contextual understanding. FastText~\citep{bojanowski2017enrichingwordvectorssubword} built upon them and improved performance by incorporating subword information. Transformer~\citep{vaswani2023attentionneed} models like BERT~\citep{devlin2019bertpretrainingdeepbidirectional} and GPT~\citep{radford2018improving} then revolutionized the field with context-aware embeddings.
Multilingual models like mBERT, XLM~\citep{lample2019crosslinguallanguagemodelpretraining}, and XLM-RoBERTa~\citep{conneau2020unsupervisedcrosslingualrepresentationlearning} further advanced cross-lingual understanding, with recent open-source LLMs pushing performance even higher~\citep{llama3, mistral_small3_blog}. Using such Transformer models, documents and representative samples can be mapped into a shared embedding space to estimate their similarity. Focusing on transparency, simplicity and efficiency in our work, we use FastText and XLM-RoBERTa for our model-based filtering.

\textbf{Multilingual Evaluation.}
Evaluating LLMs requires diverse benchmarks testing linguistic and cognitive abilities like reading comprehension, reasoning, and knowledge.
While established benchmarks such as MMLU~\citep{hendrycks2020measuring} and ARC~\citep{clark2018thinksolvedquestionanswering} exist for English evaluation, assessments in other languages often rely on translations from English sources, as seen in XNLI~\citep{conneau-etal-2018-xnli} and the machine-translated version of MMLU~\citep{lai-etal-2023-okapi}.
However, translations can be problematic, failing to capture cultural nuances or introducing "translationese"~\citep{romanou2024include}.
Recent work by~\citet{romanou2024include} and~\citet{singh2024globalmmluunderstandingaddressing} emphasizes the importance of culturally sensitive, natively collected benchmarks. Task difficulty and formulation also impact model performance when trained for shorter durations~\citep{kydlicek2024finetasksmultilingualtasks}. In our work, we follow FineTasks, a recent evaluation tasks suite by~\citet{kydlicek2024finetasksmultilingualtasks} to assess our model-based filtering approaches across multiple languages.

\section{Methods}\label{sec:method}
In this work, we present our model-based filtering approaches. Our methodology is structured into two key components: 1) we select suitable training datasets, aiming to identifying a diverse set of structured and knowledge-rich samples and 2) we describe the different models, namely FastText and Transformer embedding-based filters, used to capture and leverage these characteristics.

\subsection{Classifier Training Dataset}
\textbf{Representative Sample Selection.} Our goal is to identify a diverse set of structured and knowledge-rich samples, especially within a multilingual context. We define two criteria for our training datasets: 1) the samples must be informative and well-structured and 2) the datasets must be available in multiple languages. While some multilingual benchmark datasets meet these criteria precisely, it is important to note that we do not train the LLM directly on this data. Instead, we train a proxy model to assess pretraining data quality. Nevertheless, we must remain cautious about potentially increased pretraining data contamination stemming from this approach, as discussed in Section~\ref{sec:decontamination}. 

Based on our criteria, we selected the following datasets as representative examples.
\begin{itemize}[itemsep=1pt,topsep=0pt]
\item \textbf{\emph{Aya Collection}.} A prompt completion dataset comprising $\sim$514M samples covering a variety of tasks, generated using instruction-style templates in 101 languages~\citep{singh2024aya}. %
\item \textbf{\emph{Aya Dataset}.} Human-annotated instruction fine-tuning dataset consisting of $\sim$202K prompt-completion pairs in 65 languages~\citep{singh2024aya}. %
\item \textbf{\emph{MMLU}.} Dataset contains $\sim$14K multiple-choice knowledge questions on various topics in English~\citep{hendrycks2020measuring}. Multilingual version was translated into 14 languages by professional translators~\citep{openai2024mmmlu}. %
\item \textbf{\emph{OpenAssistant-2}.} The dataset contains $\sim$14K user-assistant conversations with multiple messages in 28 languages~\citep{fischer2024open}. %
\item \textbf{\emph{Include-Base-44}.} Multiple-choice questions focused on general and regional knowledge, extracted from academic and professional exams. Spanning 44 languages, it includes a total of $\sim$23K samples~\citep{romanou2024include}. %
\end{itemize}

\textbf{Representative Sample Collection.} 
\emph{MMLU} and \emph{Include-Base-44} are highly curated benchmark datasets, containing structured, knowledge-rich samples. The \emph{Aya Dataset} is human-curated, while \emph{OpenAssistant-2} is partially human-curated and partially generated by large language models (LLMs). In contrast, the \emph{Aya Collection} consists of various AI-generated samples without quality guarantee, though it represents the largest and most multilingual of the five.  

To address the quality difference, we create two \emph{Multilingual Knowledge Collection (MKC)} configurations which allow us to evaluate the trade-off between data quality and scale:
\begin{itemize}[itemsep=1pt,topsep=0pt]
    \item \textbf{\emph{MKC}}: Includes \emph{Include-Base-44}, \emph{OpenAssistant-2}, \emph{MMLU}, and the \emph{Aya Dataset}
    \item \textbf{\emph{MKC$^+$}}: Includes \emph{MKC} and the \emph{Aya Collection}
\end{itemize}

\textbf{Dataset Creation.}
For our model-based filtering approaches, our goal is to identify documents from the pretraining dataset that are most similar to our representative samples, with the notion of similarity determined by the specific classifier used. We can directly measure similarity to our training data, for example, by calculating cosine similarity with training samples in the embedding space. Alternatively, following the approach of \citet{li2024datacomp}, the task can be framed as a binary classification problem, with the representative samples as the positive class. For the negative class, we can subsample documents from our pretraining dataset, under the assumption that the majority of these documents are not well-structured or knowledge-rich. We use both approaches for our classifiers.

To create the binary classification training dataset, we selected up to 80K positive samples by using all examples from the smaller source datasets (e.g., \emph{Include-Base-44}) and randomly subsampling from the \emph{Aya Collection} for \emph{MKC$^+$}. The positive samples were combined with the same number of randomly sampled negative examples from FineWeb-2. The same training dataset was utilized across all model-based filtering approaches, disregarding negative samples when unnecessary. Additionally, we created a training dataset for each language individually to avoid leaking language-specific biases to data of other languages.

\textbf{Sample Pre-processing.}
We applied no pre-processing to the FineWeb-2 (negative) samples but performed minimal pre-processing on the representative (positive) samples. For instance, in datasets like \emph{MMLU} or \emph{OpenAssistant-2}, we concatenated various sample components. For the \emph{Aya Collection}, we resolved encoding issues in non-Latin languages and removed samples containing \textit{\textless unk\textgreater} tokens, which were particularly prevalent in Arabic data (37.1\%).

\subsection{FastText-based Filtering (FT)}
To efficiently process datasets with over 100 million documents~\citep{penedo2024fineweb-2}, similar to DCLM~\citep{li2024datacomp}, we used a binary FastText classifier~\citep{joulin2017bag}. FastText runs on CPU and can be deployed across multiple cores, for example using DataTrove~\citep{penedo2024datatrove}.

We trained our FastText classifier on the processed training set using 2-gram features (4-gram for Chinese). These classifiers were then used to assign scores to all documents in the pretraining dataset. To filter the dataset, we applied a score threshold based on the desired retention percentage of documents. This approach balances dataset size and the predicted quality of the samples.

\subsection{Transformer Embedding-based Filtering}
To leverage rich semantic information based on contextual relationships, we utilized Transformer model embeddings. Specifically, we selected a pretrained XLM-RoBERTa base model~\citep{conneau2020unsupervisedcrosslingualrepresentationlearning} due to its support of 100 languages, a relatively small size of 279M parameters, and its transparent training procedure. This choice enabled us to process web-scale data efficiently without being restricted to a single language and aligned with our commitment to open science.

To retain general embeddings that can be reused across methods, we opted against fine-tuning the model. For each document from our datasets, we computed the 768-dimensional embedding by mean pooling the embeddings of the output sequence. Since the model has a fixed maximum sequence length of 512 tokens, we considered only the first 512 tokens of each document, assuming they are representative of the entire document.

After computing the embeddings of our corpora, we experimented with two methods: 1) classification of embeddings using a multi-layer perceptron and 2) cosine similarity between the embeddings. As in the FastText approach, we scored each document and applied a threshold to retain the desired percentage of the highest-scoring documents.

\textbf{Multi-Layer Perceptron (MLP).} 
We trained a single-hidden-layer neural network with a dimension of 256, the ReLU activation function, a 20\% dropout, and the sigmoid function on the output. The network was trained for 6 epochs using the AdamW optimizer~\citep{loshchilov2017decoupled} with a constant learning rate $0.0003$ and binary cross-entropy loss. We computed document scores using the output layer of the MLP model, which used XML-RoBERTa document embeddings as input.

\textbf{Cosine Similarity (CS).}
We computed the document scores as the maximum cosine similarity between its embeddings and a set of $K$ randomly sampled positive sample embeddings. We experimented with varying values of $K$, including 1024, 2048, 4096, 8192, and 16384. However, we did not observe a significant differences in the documents with high scores across these variations when manually inspecting the data. To strike a balance between the diversity of the positive samples and computational efficiency, we chose $K = 8192$ for our experiments.

\section{Experiments}\label{sec:experiments}
\subsection{Experimental Setup}
\textbf{Technical Details.} We evaluate 1B-parameter Llama models~\citep{llama3} to demonstrate the effectiveness of our model-based filtering approaches. The models are trained on either 70B or 119B tokens, balancing token quality and diversity. The smaller dataset (70B tokens) exposes the model to each token at most once (with a few exceptions where some tokens appear twice). The larger dataset (119B tokens) simulates longer training, resulting in increased token repetition. Training utilizes the HuggingFace Nanotron library~\citep{nanotron} with the AdamW optimizer~\citep{loshchilov2017decoupled} and a WSD learning rate schedule~\citep{hagele2024scaling}.

To minimize the need for costly hyperparameter tuning, we maintain a consistent setup across all experiments. Specifically, we adopt the DeepSeek scaling law~\citep{deepseekai2024deepseekllmscalingopensource} with a batch size of 1.6M tokens, learning rate of 0.0008, and 2000 warmup steps.

As the base dataset, we use FineWeb-2~\citep{penedo2024fineweb-2}, which has been shown to provide a strong baseline across a variety of languages. Since FineWeb-2 is globally deduplicated, we rehydrate both filtered and unfiltered data using the hyperparameters recommended by~\citet{penedo2024fineweb-2}.

To validate our method on English, we use three datasets: FineWeb~\citep{penedo2024fineweb} as the baseline, along with FineWeb-Edu~\citep{penedo2024fineweb} and DCLM~\citep{li2024datacomp}, both of which represent the current state-of-the-art. Tokenization is performed using the multilingual Mistral v3 (Tekken) tokenizer~\citep{tekkenV3}.

\textbf{Evaluation.}
Our evaluation prioritizes a diverse range of tasks to ensure the models retain well-rounded capabilities, rather than focusing exclusively on knowledge-based tasks. Specifically, we include tasks covering reading comprehension, general knowledge, natural language understanding, common-sense reasoning, and generative tasks in the target language. To evaluate our approach, we use the HuggingFace LightEval library~\citep{lighteval}.

For French, Chinese, and Arabic, we utilize the FineTasks~\citep{kydlicek2024finetasksmultilingualtasks} multilingual evaluation suite, which is designed to provide meaningful signals even for models trained in the order of 100B tokens. We select analogous tasks for German and Danish.
For English, we rely on the SmolLM tasks suite~\citep{smollm}.
A complete list of tasks and their evaluation metrics for each language is provided in Appendix \ref{app:benchmarks}.

\textbf{Model Selection.} We follow the approach used in FineTasks~\citep{kydlicek2024finetasksmultilingualtasks} for filter selection, computing a global rank score across individual metrics and languages to determine the optimal approach. For a detailed description of the average rank computation, please refer to Appendix~\ref{app:average_rank_comp}.

\textbf{Computational Cost.} We run our experiments on a compute cluster containing four GH200 chips per node, with each GH200 chip containing 72 CPU cores and one H100 GPU. Model training on 119B tokens costs approximately 1.1K H100 compute hours, while the embedding computation of all data for 20 languages costs approximately 4K H100 compute hours. We release the embeddings publicly so this computation does not have to be repeated\footnote{\href{https://huggingface.co/datasets/epfml/FineWeb2-HQ}{huggingface.co/datasets/epfml/FineWeb2-embedded}}. In total, we use approximately 152K H100 compute hours for running our experiments and generating the document embeddings. Data selection classifier training (i.e. FastText and MLP) takes a few minutes on a CPU. Filtering using any of the approaches (i.e., FastText, MLP, or CS) is computationally inexpensive, parallelizable, and is run on CPU. Overall, filtering German, Chinese, French, Arabic, and Danish data for one filtering approach in the ablations costs approximately 60 CPU hours, which is distributed over multiple CPU cores depending on the dataset size in each language to have filtering finish in approximately 30 minutes.

\subsection{Experimental Results \& Discussion}
\subsubsection{Model Selection}\label{sec:model_selection}

    \begin{wraptable}{r}{0.5\linewidth}
        \vspace{-5.85em}
        \centering
        \caption{
        Benchmark performance comparison: Average rank between FineWeb-2 baseline and our proposed filtering methods (\emph{FT}, \emph{MLP}, and \emph{CS}) trained on \emph{MKC$^+$} or \emph{MKC}, retaining top 10\% for Chinese, German, and French, 56\% for Arabic, and 65\% for Danish. The average rank is computed across FineTasks for 1B-parameter models evaluated after 70B and 119B tokens.
        }
        \label{tab:ranking_threshold_10}
        \begin{center}
        \scalebox{1}{
        \begin{tabular}{lr}
        \toprule
        Approach &  Average Rank\\
        \midrule
        \rowcolor{Best} \emph{MLP MKC$^+$} &  4.35 \\
        \emph{MLP MKC} &  6.11 \\
        \emph{FT MKC$^+$} &  7.17 \\
        \emph{FT MKC} &  8.04 \\
        \emph{CS MKC} &  8.10 \\
        \rowcolor{Baseline} Baseline &  8.72 \\
        \emph{CS MKC$^+$} &  8.79 \\
        \bottomrule
        \end{tabular}
        }
        \end{center}
        \vspace{-1em}
    \end{wraptable}

In Section~\ref{sec:method}, we introduced several model-based filtering approaches. \textit{But which of these performs the best?} We evaluate which combination of our defined classifier training datasets (\emph{MKC} or \emph{MKC$^+$}) and filtering methods (\emph{FT}, \emph{MLP} or \emph{CS}) achieve the highest performance. Table~\ref{tab:ranking_threshold_10} presents the overall ranking across our representative language selection (Chinese, German, French, Arabic, Danish) and training runs of 70B and 119B tokens. Analogous to the DCLM filtering recipe~\citep{li2024datacomp}, the results are based on a dataset that retains 10\% of the documents for the high-resource datasets (Chinese, German, French) and keeps 56\% and 65\% of the documents for the lower-resource languages (Arabic and Danish, respectively). These percentages maintain approximately 70B tokens, under the assumption of uniform token distribution across documents. We also exclude approaches that use \emph{MKC} for training on Danish, as it lacks sufficient training data. For detailed, per-language results, please refer to Appendix \ref{app:model_selection}.

\begin{figure*}[!htb]
    \centering
    \subfigure[English (MMLU)]{\includegraphics[width=0.32\textwidth]{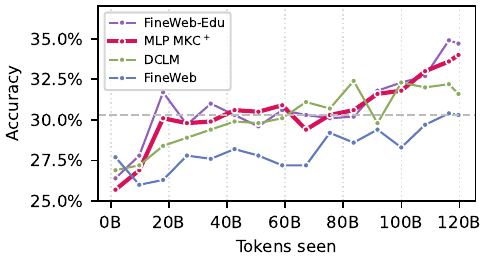}}
    \subfigure[Chinese (CMMLU)]{\includegraphics[width=0.32\textwidth]{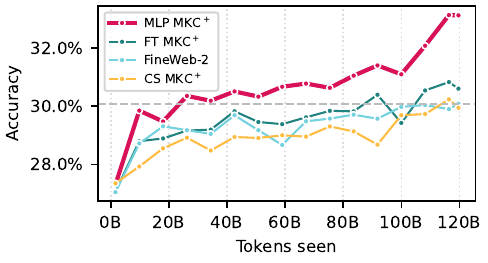}} 
    \subfigure[German (MMLU)]{\includegraphics[width=0.32\textwidth]{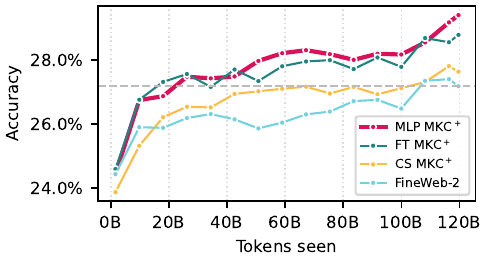}}
    \subfigure[French (MMLU)]{\includegraphics[width=0.32\textwidth]{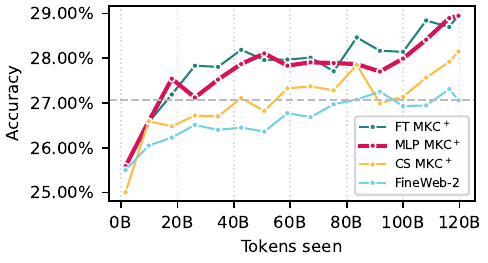}}
    \subfigure[Arabic (MMLU)]{\includegraphics[width=0.32\textwidth]{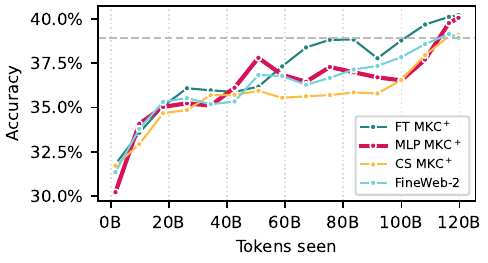}} 
    \subfigure[Danish (ARC)]{\includegraphics[width=0.32\textwidth]{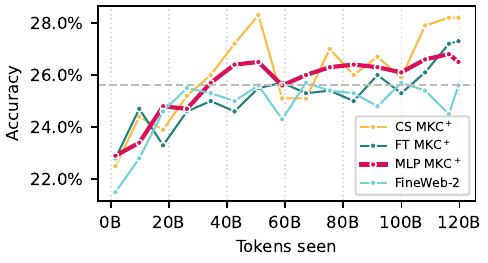}}
    \caption{
    Benchmark performance comparison: Accuracy during 119B token training between baseline methods (FineWeb, DCLM, FineWeb-Edu, FineWeb-2) and our proposed filtering approaches (\emph{FT}, \emph{MLP}, and \emph{CS}), trained on \emph{MKC$^+$}. Our approaches use 10\% data retention for English, Chinese, German, and French, 56\% for Arabic, and 65\% for Danish. For English, Chinese, German, and French, baseline-level performance is reached at approximately 20B tokens (16.7\% of total).
    }
    \label{fig:mmlu_plot}
\end{figure*}

Table~\ref{tab:ranking_threshold_10} demonstrates that \emph{MLP MKC$^+$} approach outperforms all other approaches. Interestingly, the high- and low-scored samples presented in Appendix~\ref{app:examples} align with the observed rankings. Figure~\ref{fig:mmlu_plot} further highlights the strong performance of \emph{MLP MKC$^+$}, particularly for high-resource languages, where it largely outperforms the baseline. For lower-resource languages—where less data was filtered—the performance gains are less pronounced. Notably, \emph{FT} filtering is also competitive. Given the computational expense of XLM-RoBERTa embeddings, FastText can be a promising alternative in resource-constrained setups.

\subsubsection{Threshold Selection}\label{sec:threshold_selection}
    \begin{wraptable}{r}{0.5\linewidth}
        \vspace{-4.05em}
        \centering
        \caption{
        Benchmark performance comparison: Average rank between FineWeb-2 baseline and our proposed filtering methods (\emph{FT}, \emph{MLP}) trained on \emph{MKC$^+$} or \emph{MKC}, retaining top 10\%, 15\% or 20\% of documents. The average rank is computed across FineTasks for 1B-parameter models evaluated on Chinese, German and French after 70B and 119B tokens.        
        }
        \label{tab:ranking_threshold_10_15_20}
        \begin{center}
        \scalebox{0.8}{
        \begin{tabular}{lcr}
        \toprule
        Approach & Threshold & Average Rank \\
        \midrule
        \rowcolor{Best} \emph{MLP MKC$^+$} & 10\% & 8.85 \\
        \emph{MLP MKC$^+$} & 15\% & 9.44 \\
        \emph{MLP MKC} & 20\% & 11.37 \\
        \emph{MLP MKC} & 15\% & 11.70 \\
        \emph{MLP MKC} & 10\% & 11.95 \\
        \emph{MLP MKC$^+$} & 20\% & 11.97 \\
        \emph{FT MKC$^+$} & 10\% & 13.92 \\
        \emph{FT MKC} & 15\% & 14.62 \\
        \emph{FT MKC} & 10\% & 14.74 \\
        \emph{FT MKC} & 20\% & 15.62 \\
        \emph{FT MKC$^+$} & 15\% & 16.27 \\
        \emph{FT MKC$^+$} & 20\% & 16.51 \\
        \rowcolor{Baseline} Baseline & -- & 18.55 \\
        \bottomrule
        \end{tabular}
        }
        \end{center}
        \vspace{-1em}
    \end{wraptable}

In Section~\ref{sec:model_selection}, we base our model selection on experiments that retain top 10\% of the data for high-resource languages. \textit{But is this the optimal threshold?} Following the methodology of \citet{li2024datacomp}, we analyze the impact of varying filter strengths on performance for Chinese, German, and French, using our \emph{MLP} and \emph{FT} filtering methods. The results are summarized in Table~\ref{tab:ranking_threshold_10_15_20}, with a comprehensive analysis, including results for \emph{CS}, provided in Appendix~\ref{app:threshold_selection} (Table~\ref{tab:ranking_threshold_15_20_all}). Consistent with their findings, we observe that retaining top 10\% of the data is a competitive threshold, particularly for approaches using the \emph{MKC$^+$} dataset. Interestingly, approaches using \emph{MKC} perform better with higher retention. In Appendix~\ref{app:threshold_selection}, we investigate how some filters' bias toward shorter documents affects threshold selection, though our analysis indicates multiple factors contribute to optimal threshold determination.

\subsubsection{Training Data Analysis}

The experiments in Sections~\ref{sec:model_selection} and~\ref{sec:threshold_selection} are based on the training datasets \emph{MKC} and \emph{MKC$^+$}. \textit{But is the diversity introduced by combining various base datasets truly necessary?} We evaluate the impact of each base dataset individually and compare it to the combined \emph{MKC$^+$} dataset.
For this ablation study, we use our best filtering method (\emph{MLP} with a top 10\% retention) and train the models on 30B tokens. This token count is chosen to match the size of the smallest filtered dataset, ensuring consistency across comparisons.

The results, presented in Table~\ref{tab:ranking_training_data}, show that despite the absence of a quality guarantee for all samples in the \emph{Aya Collection}, this dataset yields strong performance, making our approach applicable for various languages. Overall, we observe that the diversity resulting from combining all individual training datasets gives the best results.

    \begin{wraptable}{r}{0.5\linewidth}
        \vspace{-2em}
        \centering
        \caption{
        Benchmark performance comparison: Average rank between FineWeb-2 baseline and \emph{MLP} filtering trained on either full \emph{MKC$^+$} or its individual components, retaining top 10\% for Chinese, German, and French, 56\% for Arabic, and 65\% for Danish. The average rank is computed across FineTasks for 1B-parameter models trained on 30B tokens per language.
        }
        \label{tab:ranking_training_data}
        \begin{center}
        \scalebox{0.8}{
        \begin{tabular}{lr}
        \toprule
        Dataset & Average Rank \\
        \midrule
        \rowcolor{Best}\emph{MKC$^+$} & 2.52 \\
        \emph{Aya Collection} & 2.91 \\
        \emph{Aya Dataset} & 3.17 \\
        \emph{MMLU} & 3.57 \\
        \rowcolor{Baseline} Baseline & 4.09 \\
        \emph{OpenAssistant-2} & 4.53 \\
        \emph{Include-Base-44} & 5.42 \\
        \bottomrule
        \end{tabular}
        }
        \end{center}
        \vspace{-1em}
    \end{wraptable}

Interestingly, models trained exclusively on \emph{Include-Base-44} and \emph{OpenAssistant-2} perform worse overall than the baseline.
This may reflect dataset characteristics—\emph{Include-Base-44} is small and domain-specific, containing mostly driving license exam questions in its German subset. \emph{OpenAssistant-2} includes a limited number of samples, with fewer than 2K positive samples per training set, which likely negatively impacts classifier performance.
In Appendix~\ref{app:training_data_analysis}, we reexamine how document length bias relates to model performance, confirming our Section~\ref{sec:threshold_selection} finding that performance depends on factors beyond document length.
In Appendix~\ref{app:sec:replay}, we further verify our filtering approach preserves sufficient dataset diversity.

\subsubsection{Approach Validation on English}\label{sec:eng_validation}
    \begin{wraptable}{r}{0.5\linewidth}
        \vspace{-2em}
        \centering
        \caption{English benchmark performance: Our \emph{MLP MKC$^+$} approach (top 10\% documents) compared to FineWeb, DCLM, and FineWeb-Edu baselines.
        The average rank is computed across SmolLM tasks using 1B-parameter models trained on 119B tokens.
        }
        \label{tab:ranking_english}
        \begin{center}
        \resizebox{0.99\linewidth}{!}{
        \begin{tabular}{l>{\columncolor{Best}}lll>{\columncolor{Baseline}}r}
        \toprule
        Dataset & Ours & DCLM$^*$ & FW-Edu$^*$ & FW$^*$ \\
        \midrule
        \rowcolor{AverageRank} Average Rank & 1.8333 & 2.3889 & 2.4444 & 3.3333 \\
        ARC (Challenge) & 0.3550 & 0.3530 & \textbf{0.3850} & 0.3010 \\
        ARC (Easy) & 0.6670 & 0.6470 & \textbf{0.6970} & 0.5880 \\
        CommonsenseQA & 0.3870 & \textbf{0.4100} & 0.3770 & 0.3850 \\
        HellaSwag & \textbf{0.6040} & 0.5960 & 0.5700 & 0.5930 \\
        MMLU & 0.3400 & 0.3160 & \textbf{0.3470} & 0.3030 \\
        OpenBookQA & 0.3860 & 0.3840 & \textbf{0.4180} & 0.3560 \\
        PIQA & 0.7510 & 0.7510 & 0.7410 & \textbf{0.7620} \\
        WinoGrande & \textbf{0.5720} & 0.5610 & 0.5660 & 0.5550 \\
        TriviaQA & 0.0820 & \textbf{0.1240} & 0.0320 & 0.0370 \\
        \bottomrule
        \end{tabular}}
        \end{center}
        \vspace{-1.5em}
    \end{wraptable}
Previous experiments have shown strong performance of our \emph{MLP MKC$^+$} approach. \textit{But do these results translate to English?}
Table~\ref{tab:ranking_english} presents the performance of \emph{MLP MKC$^+$} with 10\% retention applied to the English FineWeb dataset~\cite{penedo2024fineweb}.
Our method is compared against FineWeb and baselines using model-based filtered datasets, including DCLM~\cite{li2024datacomp} and FineWeb-Edu~\citep{penedo2024fineweb}. To save computational resources, we use the 6 most recent FineWeb and FineWeb-Edu dumps and the first partition of DCLM\footnote{\href{https://huggingface.co/datasets/mlfoundations/dclm-baseline-1.0-parquet}{huggingface.co/datasets/mlfoundations/dclm-baseline-1.0-parquet}}, which we denote with $^*$. Each of these subsets contains more than 119B tokens, with FineWeb retaining this size even after applying our filtering retaining top 10\% of the documents.

While each approach demonstrates strengths in different benchmarks, as seen from Table~\ref{tab:ranking_english} and Figure~\ref{fig:mmlu_plot}, the overall average rank results indicate that our method outperforms all other baselines.

\subsubsection{Data Contamination Analysis}\label{sec:decontamination}
Our LLMs are never trained on benchmark datasets. \textit{But is the strong performance observed in the previous sections primarily due to an increased ratio of data contamination?}
To ensure the validity of our approach, we conduct decontamination experiments, as web crawl data may include evaluation benchmark tasks. While~\citet{li2024datacomp} addressed similar concerns, our approach follows the methodology of~\citet{brown2020language}. Specifically, we perform 13-gram decontamination of the LLM training data separately for English and French evaluation benchmarks. However, unlike the original approach, we remove the entire document if it is flagged as contaminated, using the implementation provided in DataTrove~\citep{penedo2024datatrove}.

Tables \ref{tab:ranking_decont_eng_Latn_119} and \ref{tab:ranking_decont_fra_Latn_119} present the results of decontamination experiments for English and French, respectively. We used the following experimental setup (removed document contamination rates): baseline FineWeb English (0.16\%), \emph{MLP MKC$^+$} English with 10\% retention (0.19\%), baseline FineWeb-2 French (0.14\%), and \emph{MLP MKC$^+$} French with 10\% retention (0.14\%). All models were trained on 119B tokens. Additionally, we compare the results against equivalent training runs without decontamination to further analyze its impact. For an example of a contaminated sample, see Appendix~\ref{app:decontmination}.

For English models, decontamination slightly reduces performance both for our approach and baseline FineWeb data. Even after decontamination, our approach still outperforms the baseline trained on non-decontaminated data. For French models, our approach performs similarly on decontaminated and non-decontaminated data, both outperforming baseline FineWeb-2. Interestingly, decontaminated baseline data yields better results than its non-decontaminated counterpart.

\begin{table}[!htb]\small
    \centering
    \begin{minipage}[t]{0.48\textwidth}
        \centering
        \caption{
        English benchmark performance: Our \emph{MLP MKC$^+$} approach (retaining top 10\% documents) in both decontaminated ($D$) and non-decontaminated versions, compared to baseline FineWeb datasets with the same variants. The average rank is computed across SmolLM tasks for 1B-parameter models trained on 119B tokens.
        }
        \label{tab:ranking_decont_eng_Latn_119}
        \begin{center}
        \resizebox{0.99\linewidth}{!}{
        \begin{tabular}{l>{\columncolor{Best}}ll>{\columncolor{Baseline}}lr}
        \toprule
        Dataset & Ours & Ours$_D$ & FW$^*$ & FW$^*_D$ \\
        \midrule
        \rowcolor{AverageRank} Average Rank & 1.5000 & 2.1111 & 3.0556 & 3.3333 \\
        ARC (Challenge) & \textbf{0.3550} & 0.3440 & 0.3010 & 0.2880 \\
        ARC (Easy) & \textbf{0.6670} & 0.6520 & 0.5880 & 0.5700 \\
        CommonsenseQA & 0.3870 & \textbf{0.4000} & 0.3850 & 0.3820 \\
        HellaSwag & \textbf{0.6040} & \textbf{0.6040} & 0.5930 & 0.5890 \\
        MMLU & \textbf{0.3400} & 0.3220 & 0.3030 & 0.3050 \\
        OpenBookQA & \textbf{0.3860} & 0.3840 & 0.3560 & 0.3740 \\
        PIQA & 0.7510 & 0.7590 & \textbf{0.7620} & 0.7600 \\
        WinoGrande & \textbf{0.5720} & 0.5550 & 0.5550 & 0.5570 \\
        TriviaQA  & \textbf{0.0820} & 0.0380 & 0.0370 & 0.0250 \\
        \bottomrule
        \end{tabular}}
        \end{center}
    \end{minipage}
        \hfill 
    \begin{minipage}[t]{0.48\textwidth}
        \centering
            \caption{French Benchmark performance: Our \emph{MLP MKC$^+$} approach (retaining top 10\% of the documents) in both decontaminated ($D$) and non-decontaminated versions, compared to baseline FineWeb-2 datasets with the same variants. The average rank is computed across FineTasks for 1B-parameter models trained on 119B tokens.
            }
        \label{tab:ranking_decont_fra_Latn_119}
        \begin{center}
        \resizebox{0.99\linewidth}{!}{
        \begin{tabular}{l>{\columncolor{Best}}lll>{\columncolor{Baseline}}r}
        \toprule
        Dataset & Ours & Ours$_D$ & FW-2$_D$ & FW-2 \\
        \midrule
        \rowcolor{AverageRank} Average Rank & 2.0556 & 2.0556 & 2.7222 & 3.1667 \\
        Belebele & 0.3533 & 0.3400 & \textbf{0.3778} & 0.3444 \\
        HellaSwag & \textbf{0.5380} & 0.5350 & 0.5180 & 0.5180 \\
        X-CSQA & 0.2740 & 0.2810 & 0.2730 & \textbf{0.2870} \\
        XNLI 2.0 & \textbf{0.7400} & \textbf{0.7400} & 0.7070 & 0.7180 \\
        FQuAD  & 0.2803 & 0.2620 & \textbf{0.2890} & 0.2401 \\
        MMLU & \textbf{0.2895} & 0.2875 & 0.2711 & 0.2706 \\
        Mintaka  & 0.0438 & \textbf{0.0797} & 0.0658 & 0.0712 \\
        X-CODAH & 0.2667 & \textbf{0.2900} & 0.2800 & 0.2633 \\
        ARC (Challenge) & \textbf{0.3180} & 0.3110 & 0.2880 & 0.2850 \\
        \bottomrule
        \end{tabular}}
        \end{center}
    \end{minipage}
\end{table}

\subsubsection{Impact on Multilingual Training - Mitigating the Curse of Multilinguality}\label{sec:multilingual_llm}
    \begin{wraptable}{r}{0.5\linewidth}
        \vspace{-2em}
        \centering
        \caption{
        French benchmark performance: Multilingual LLMs ($M$) trained on FineWeb-2 or our \emph{MLP MKC$^+$} refined dataset (retaining top 10\% for Chinese, German and French, 56\% for Arabic, 65\% for Danish) with 595B tokens, compared to monolingual models trained on 119B tokens. The average rank is computed across FineTasks for 1B-parameter models.
        }
        \label{tab:ranking_multilingual_fra_Latn_595}
        \begin{center}
        \resizebox{0.99\linewidth}{!}{
        \begin{tabular}{ll>{\columncolor{Best}}l>{\columncolor{Baseline}}lr}
        \toprule
        Dataset & Ours$_M$ & Ours & FW-2 & FW-2$_M$ \\
        \midrule
        \rowcolor{AverageRank} Average Rank & 1.8333 & 2.0556 & 3.0000 & 3.1111 \\
        Belebele & \textbf{0.3667} & 0.3533 & 0.3444 & 0.3511 \\
        HellaSwag & 0.5270 & \textbf{0.5380} & 0.5180 & 0.4970 \\
        X-CSQA & 0.2740 & 0.2740 & \textbf{0.2870} & 0.2750 \\
        XNLI 2.0 & \textbf{0.7660} & 0.7400 & 0.7180 & 0.7330 \\
        FQuAD  & \textbf{0.3212} & 0.2803 & 0.2401 & 0.2459 \\
        MMLU & 0.2841 & \textbf{0.2895} & 0.2706 & 0.2735 \\
        Mintaka  & 0.0456 & 0.0438 & \textbf{0.0712} & 0.0579 \\
        X-CODAH & \textbf{0.2900} & 0.2667 & 0.2633 & 0.2567 \\
        ARC (Challenge) & 0.2970 & \textbf{0.3180} & 0.2850 & 0.2670 \\
        \bottomrule
        \end{tabular}}
        \end{center}
        \vspace{-2em}
    \end{wraptable}
Although not our main focus, we found that our refined datasets boost the performance of multilingual models%
. We trained a multilingual 1B-parameter model on 595B tokens (119B per language), covering all five languages: Chinese, German, French, Arabic and Danish. We compared each language's results to its monolingual counterpart trained on 119B tokens. Training is performed once for our filtered data and once for original (unfiltered) FineWeb-2. 

The results for French are presented in Table~\ref{tab:ranking_multilingual_fra_Latn_595}. Surprisingly, the \emph{curse} of multilinguality~\citep{chang2023multilingualitycurselanguagemodeling}
turns into a \emph{benefit} for our quality filtered datasets: The multilingual model outperforms its monolingual counterpart, when both models have seen an equal amount of tokens of the language of interest.
Meanwhile, for unfiltered training data, the multilingual LLM suffers from the \emph{curse} as expected.
The disappearance of the \emph{curse} is consistent across all languages except Chinese. 
Detailed results for the other languages are provided in Appendix~\ref{app:multilingual}.

\section{Conclusion}
In this work, we developed a framework for model-based filtering of web-scale multilingual pretraining datasets, demonstrating consistent improvements on LLM benchmarks across a wide range of languages. Our Transformer embedding-based classifier, \emph{MLP MKC$^+$}, outperforms state-of-the-art methods on both English and multilingual datasets, even when decontaminating the datasets or using them for training multilingual LLMs. While our FastText-based filtering approach performed well and shows promise in resource-constrained setups, \emph{MLP MKC$^+$} consistently outperformed all other methods and can be easily scaled to other languages. These results provide strong empirical evidence supporting our expansion of the framework to 20 languages. We release the corresponding refined pretraining datasets and code, contributing to the advancement of multilingual language modeling.

\newpage
\begin{ack}
We thank Guilherme Penedo, Hynek Kydlíček, and Leandro von Werra for their help with FineWeb-2 data, and to Alex Hägele for providing feedback on the paper draft. 

This work was supported as part of the Swiss AI Initiative by a grant from the Swiss National Supercomputing Centre (CSCS) under project ID a06 on Alps.
\end{ack}

\newpage
\bibliographystyle{abbrvnat}
\bibliography{references}

@article{li2024datacomp,
  title={{D}ata{C}omp-{LM}: In search of the next generation of training sets for language models},
  author={Li, Jeffrey and Fang, Alex and Smyrnis, Georgios and Ivgi, Maor and Jordan, Matt and Gadre, Samir and Bansal, Hritik and Guha, Etash and Keh, Sedrick and Arora, Kushal and others},
  journal={arXiv preprint arXiv:2406.11794},
  year={2024}
}

@article{penedo2024fineweb,
  title={{T}he {F}ine{W}eb {D}atasets: {D}ecanting the {W}eb for the {F}inest {T}ext {D}ata at {S}cale},
  author={Penedo, Guilherme and Kydl{\'\i}{\v{c}}ek, Hynek and Lozhkov, Anton and Mitchell, Margaret and Raffel, Colin and Von Werra, Leandro and Wolf, Thomas and others},
  journal={arXiv preprint arXiv:2406.17557},
  year={2024}
}

@software{penedo2024fineweb-2,
  author = {Penedo, Guilherme and Kydlíček, Hynek and Sabolčec, Vinko and Messmer, Bettina and Foroutan, Negar and Jaggi, Martin and von Werra, Leandro and Wolf, Thomas},
  title = {{F}ine{W}eb2: {A} sparkling update with 1000s of languages},
  month = dec,
  year = 2024,
  doi = { 10.57967/hf/3744 },
  url = {https://huggingface.co/datasets/HuggingFaceFW/fineweb-2},
  note = {Accessed 30 Jan. 2025}
}

@article{fischer2024open,
  title={Open Assistant Toolkit--version 2},
  author={Fischer, Sophie and Rossetto, Federico and Gemmell, Carlos and Ramsay, Andrew and Mackie, Iain and Zubel, Philip and Tecklenburg, Niklas and Dalton, Jeffrey},
  journal={arXiv preprint arXiv:2403.00586},
  year={2024}
}

@misc{openai2024mmmlu,
  title={{MMMLU}},
  author={OpenAI},
  year={2024},
  url={https://huggingface.co/datasets/openai/MMMLU},
  note = {Accessed 30 Jan. 2025}
}

@misc{singh2024aya,
      title={Aya Dataset: An Open-Access Collection for Multilingual Instruction Tuning}, 
      author={Shivalika Singh and Freddie Vargus and Daniel Dsouza and Börje F. Karlsson and Abinaya Mahendiran and Wei-Yin Ko and Herumb Shandilya and Jay Patel and Deividas Mataciunas and Laura OMahony and Mike Zhang and Ramith Hettiarachchi and Joseph Wilson and Marina Machado and Luisa Souza Moura and Dominik Krzemiński and Hakimeh Fadaei and Irem Ergün and Ifeoma Okoh and Aisha Alaagib and Oshan Mudannayake and Zaid Alyafeai and Vu Minh Chien and Sebastian Ruder and Surya Guthikonda and Emad A. Alghamdi and Sebastian Gehrmann and Niklas Muennighoff and Max Bartolo and Julia Kreutzer and Ahmet Üstün and Marzieh Fadaee and Sara Hooker},
      year={2024},
      eprint={2402.06619},
      archivePrefix={arXiv},
      primaryClass={cs.CL}
}

@article{hendrycks2020measuring,
  title={Measuring massive multitask language understanding},
  author={Hendrycks, Dan and Burns, Collin and Basart, Steven and Zou, Andy and Mazeika, Mantas and Song, Dawn and Steinhardt, Jacob},
  journal={arXiv preprint arXiv:2009.03300},
  year={2020}
}

@article{romanou2024include,
  title={Include: Evaluating multilingual language understanding with regional knowledge},
  author={Romanou, Angelika and Foroutan, Negar and Sotnikova, Anna and Chen, Zeming and Nelaturu, Sree Harsha and Singh, Shivalika and Maheshwary, Rishabh and Altomare, Micol and Haggag, Mohamed A and Amayuelas, Alfonso and others},
  journal={arXiv preprint arXiv:2411.19799},
  year={2024}
}

@InProceedings{joulin2017bag,
  title={Bag of Tricks for Efficient Text Classification},
  author={Joulin, Armand and Grave, Edouard and Bojanowski, Piotr and Mikolov, Tomas},
  booktitle={Proceedings of the 15th Conference of the European Chapter of the Association for Computational Linguistics: Volume 2, Short Papers},
  month={April},
  year={2017},
  publisher={Association for Computational Linguistics},
  pages={427--431},
}

@misc{conneau2020unsupervisedcrosslingualrepresentationlearning,
      title={Unsupervised Cross-lingual Representation Learning at Scale}, 
      author={Alexis Conneau and Kartikay Khandelwal and Naman Goyal and Vishrav Chaudhary and Guillaume Wenzek and Francisco Guzmán and Edouard Grave and Myle Ott and Luke Zettlemoyer and Veselin Stoyanov},
      year={2020},
      eprint={1911.02116},
      archivePrefix={arXiv},
      primaryClass={cs.CL},
      url={https://arxiv.org/abs/1911.02116}, 
}

@misc{kydlicek2024finetasksmultilingualtasks,
      title={{F}ine{T}asks: {F}inding signal in a haystack of 200+ multilingual tasks},
      author={Hynek Kydlíček and Guilherme Penedo and Clémentine Fourier and Nathan Habib and Thomas Wolf},
      url={https://huggingface.co/spaces/HuggingFaceFW/blogpost-fine-tasks},
      year={2024},
      note = {Accessed 30 Jan. 2025}

}

@misc{lighteval,
  author = {Fourrier, Clémentine and Habib, Nathan and Wolf, Thomas and Tunstall, Lewis},
  title = {{L}ight{E}val: {A} lightweight framework for {LLM} evaluation},
  year = {2023},
  version = {0.5.0},
  url = {https://github.com/huggingface/lighteval},
      year={2024},
}

@misc{nanotron,
      title={{N}anotron},
      author={{Hugging Face}},
      url={https://github.com/huggingface/nanotron},
      year={2024},
      note = {Accessed 30 Jan. 2025}
}

@misc{penedo2024datatrove,
  author = {Penedo, Guilherme and Kydlíček, Hynek and Cappelli, Alessandro and Sasko, Mario and Wolf, Thomas},
  title = {{D}ata{T}rove: large scale data processing},
  year = {2024},
  publisher = {GitHub},
  journal = {GitHub repository},
  url = {https://github.com/huggingface/datatrove},
  note = {Accessed 30 Jan. 2025}

}

@misc{deepseekai2024deepseekllmscalingopensource,
      title={{D}eep{S}eek {LLM}: {S}caling {O}pen-{S}ource {L}anguage {M}odels with {L}ongtermism}, 
      author={DeepSeek-AI and : and Xiao Bi and Deli Chen and Guanting Chen and Shanhuang Chen and Damai Dai and Chengqi Deng and Honghui Ding and Kai Dong and Qiushi Du and Zhe Fu and Huazuo Gao and Kaige Gao and Wenjun Gao and Ruiqi Ge and Kang Guan and Daya Guo and Jianzhong Guo and Guangbo Hao and Zhewen Hao and Ying He and Wenjie Hu and Panpan Huang and Erhang Li and Guowei Li and Jiashi Li and Yao Li and Y. K. Li and Wenfeng Liang and Fangyun Lin and A. X. Liu and Bo Liu and Wen Liu and Xiaodong Liu and Xin Liu and Yiyuan Liu and Haoyu Lu and Shanghao Lu and Fuli Luo and Shirong Ma and Xiaotao Nie and Tian Pei and Yishi Piao and Junjie Qiu and Hui Qu and Tongzheng Ren and Zehui Ren and Chong Ruan and Zhangli Sha and Zhihong Shao and Junxiao Song and Xuecheng Su and Jingxiang Sun and Yaofeng Sun and Minghui Tang and Bingxuan Wang and Peiyi Wang and Shiyu Wang and Yaohui Wang and Yongji Wang and Tong Wu and Y. Wu and Xin Xie and Zhenda Xie and Ziwei Xie and Yiliang Xiong and Hanwei Xu and R. X. Xu and Yanhong Xu and Dejian Yang and Yuxiang You and Shuiping Yu and Xingkai Yu and B. Zhang and Haowei Zhang and Lecong Zhang and Liyue Zhang and Mingchuan Zhang and Minghua Zhang and Wentao Zhang and Yichao Zhang and Chenggang Zhao and Yao Zhao and Shangyan Zhou and Shunfeng Zhou and Qihao Zhu and Yuheng Zou},
      year={2024},
      eprint={2401.02954},
      archivePrefix={arXiv},
      primaryClass={cs.CL},
      url={https://arxiv.org/abs/2401.02954}, 
}

@article{loshchilov2017decoupled,
      title={Decoupled Weight Decay Regularization}, 
      author={Ilya Loshchilov and Frank Hutter},
      year={2019},
      eprint={1711.05101},
      archivePrefix={arXiv},
      primaryClass={cs.LG},
      url={https://arxiv.org/abs/1711.05101},
}

@misc{chen2023meditron70bscalingmedicalpretraining,
      title={MEDITRON-70B: Scaling Medical Pretraining for Large Language Models}, 
      author={Zeming Chen and Alejandro Hernández Cano and Angelika Romanou and Antoine Bonnet and Kyle Matoba and Francesco Salvi and Matteo Pagliardini and Simin Fan and Andreas Köpf and Amirkeivan Mohtashami and Alexandre Sallinen and Alireza Sakhaeirad and Vinitra Swamy and Igor Krawczuk and Deniz Bayazit and Axel Marmet and Syrielle Montariol and Mary-Anne Hartley and Martin Jaggi and Antoine Bosselut},
      year={2023},
      eprint={2311.16079},
      archivePrefix={arXiv},
      primaryClass={cs.CL},
      url={https://arxiv.org/abs/2311.16079}, 
}

@misc{bethune2025scalinglawsforgettingfinetuning,
      title={Scaling Laws for Forgetting during Finetuning with Pretraining Data Injection}, 
      author={Louis Bethune and David Grangier and Dan Busbridge and Eleonora Gualdoni and Marco Cuturi and Pierre Ablin},
      year={2025},
      eprint={2502.06042},
      archivePrefix={arXiv},
      primaryClass={cs.LG},
      url={https://arxiv.org/abs/2502.06042}, 
}

@article{hagele2024scaling,
  title={Scaling Laws and Compute-Optimal Training Beyond Fixed Training Durations},
  author={H{\"a}gele, Alexander and Bakouch, Elie and Kosson, Atli and Allal, Loubna Ben and Von Werra, Leandro and Jaggi, Martin},
  journal={arXiv preprint arXiv:2405.18392},
  year={2024}
}

@misc{mistral_small3_blog,
      title={Mistral Small 3},
      author={{Mistral AI}},
      url={https://mistral.ai/news/mistral-small-3/},
      note = {Accessed 30 Jan. 2025},
      year={2025}
}

@misc{tekkenV3,
      title={v3 (Tekken) Tokenizer},
      author={{Mistral AI}},
      url={https://docs.mistral.ai/guides/tokenization/},
      note = {Accessed 30 Jan. 2025},
      year={2024}
}

@misc{smollm,
      title={{SmolLM} - blazingly fast and remarkably powerful},
      author={{Hugging Face}},
      url={https://huggingface.co/blog/smollm},
      note = {Accessed 30 Jan. 2025},
      year={2024},
}

@article{brown2020language,
  title={Language models are few-shot learners},
  author={Brown, Tom and Mann, Benjamin and Ryder, Nick and Subbiah, Melanie and Kaplan, Jared D and Dhariwal, Prafulla and Neelakantan, Arvind and Shyam, Pranav and Sastry, Girish and Askell, Amanda and others},
  journal={Advances in neural information processing systems},
  volume={33},
  pages={1877--1901},
  year={2020}
}

@misc{llama3,
      title={{The Llama 3 Herd of Models}}, 
      author="{Llama Team}",
      year={2024},
      eprint={2407.21783},
      archivePrefix={arXiv},
      primaryClass={cs.AI},
      url={https://arxiv.org/abs/2407.21783}, 
}

@article{rae2021scaling,
  title={Scaling language models: Methods, analysis \& insights from training gopher},
  author={Rae, Jack W and Borgeaud, Sebastian and Cai, Trevor and Millican, Katie and Hoffmann, Jordan and Song, Francis and Aslanides, John and Henderson, Sarah and Ring, Roman and Young, Susannah and others},
  journal={arXiv preprint arXiv:2112.11446},
  year={2021}
}

@article{de2024new,
  title={A New Massive Multilingual Dataset for High-Performance Language Technologies},
  author={De Gibert, Ona and Nail, Graeme and Arefyev, Nikolay and Ba{\~n}{\'o}n, Marta and Van Der Linde, Jelmer and Ji, Shaoxiong and Zaragoza-Bernabeu, Jaume and Aulamo, Mikko and Ram{\'\i}rez-S{\'a}nchez, Gema and Kutuzov, Andrey and others},
  journal={arXiv preprint arXiv:2403.14009},
  year={2024}
}

@article{soldaini2024dolma,
  title={Dolma: An open corpus of three trillion tokens for language model pretraining research},
  author={Soldaini, Luca and Kinney, Rodney and Bhagia, Akshita and Schwenk, Dustin and Atkinson, David and Authur, Russell and Bogin, Ben and Chandu, Khyathi and Dumas, Jennifer and Elazar, Yanai and others},
  journal={arXiv preprint arXiv:2402.00159},
  year={2024}
}

@article{laurenccon2022bigscience,
  title={The bigscience roots corpus: A 1.6 tb composite multilingual dataset},
  author={Lauren{\c{c}}on, Hugo and Saulnier, Lucile and Wang, Thomas and Akiki, Christopher and Villanova del Moral, Albert and Le Scao, Teven and Von Werra, Leandro and Mou, Chenghao and Gonz{\'a}lez Ponferrada, Eduardo and Nguyen, Huu and others},
  journal={Advances in Neural Information Processing Systems},
  volume={35},
  pages={31809--31826},
  year={2022}
}

@article{penedo2023refinedweb,
  title={{The RefinedWeb dataset for Falcon LLM: Outperforming curated corpora with web data only}},
  author={Penedo, Guilherme and Malartic, Quentin and Hesslow, Daniel and Cojocaru, Ruxandra and Alobeidli, Hamza and Cappelli, Alessandro and Pannier, Baptiste and Almazrouei, Ebtesam and Launay, Julien},
  journal={Advances in Neural Information Processing Systems},
  volume={36},
  pages={79155--79172},
  year={2023}
}

@article{wenzek2019ccnet,
  title={CCNet: Extracting high quality monolingual datasets from web crawl data},
  author={Wenzek, Guillaume and Lachaux, Marie-Anne and Conneau, Alexis and Chaudhary, Vishrav and Guzm{\'a}n, Francisco and Joulin, Armand and Grave, Edouard},
  journal={arXiv preprint arXiv:1911.00359},
  year={2019}
}

@misc{marion2023moreinvestigatingdatapruning,
      title={When Less is More: Investigating Data Pruning for Pretraining LLMs at Scale}, 
      author={Max Marion and Ahmet Üstün and Luiza Pozzobon and Alex Wang and Marzieh Fadaee and Sara Hooker},
      year={2023},
      eprint={2309.04564},
      archivePrefix={arXiv},
      primaryClass={cs.CL},
      url={https://arxiv.org/abs/2309.04564}, 
}

@misc{ankner2024perplexedperplexityperplexitybaseddata,
      title={Perplexed by Perplexity: Perplexity-Based Data Pruning With Small Reference Models}, 
      author={Zachary Ankner and Cody Blakeney and Kartik Sreenivasan and Max Marion and Matthew L. Leavitt and Mansheej Paul},
      year={2024},
      eprint={2405.20541},
      archivePrefix={arXiv},
      primaryClass={cs.LG},
      url={https://arxiv.org/abs/2405.20541}, 
}

@misc{gunasekar2023textbooksneed,
      title={Textbooks Are All You Need}, 
      author={Suriya Gunasekar and Yi Zhang and Jyoti Aneja and Caio César Teodoro Mendes and Allie Del Giorno and Sivakanth Gopi and Mojan Javaheripi and Piero Kauffmann and Gustavo de Rosa and Olli Saarikivi and Adil Salim and Shital Shah and Harkirat Singh Behl and Xin Wang and Sébastien Bubeck and Ronen Eldan and Adam Tauman Kalai and Yin Tat Lee and Yuanzhi Li},
      year={2023},
      eprint={2306.11644},
      archivePrefix={arXiv},
      primaryClass={cs.CL},
      url={https://arxiv.org/abs/2306.11644}, 
}

@misc{wettig2024quratingselectinghighqualitydata,
      title={QuRating: Selecting High-Quality Data for Training Language Models}, 
      author={Alexander Wettig and Aatmik Gupta and Saumya Malik and Danqi Chen},
      year={2024},
      eprint={2402.09739},
      archivePrefix={arXiv},
      primaryClass={cs.CL},
      url={https://arxiv.org/abs/2402.09739}, 
}

@misc{sachdeva2024traindataefficientllms,
      title={How to Train Data-Efficient LLMs}, 
      author={Noveen Sachdeva and Benjamin Coleman and Wang-Cheng Kang and Jianmo Ni and Lichan Hong and Ed H. Chi and James Caverlee and Julian McAuley and Derek Zhiyuan Cheng},
      year={2024},
      eprint={2402.09668},
      archivePrefix={arXiv},
      primaryClass={cs.LG},
      url={https://arxiv.org/abs/2402.09668}, 
}

@article{raffel2020exploring,
  title={Exploring the limits of transfer learning with a unified text-to-text transformer},
  author={Raffel, Colin and Shazeer, Noam and Roberts, Adam and Lee, Katherine and Narang, Sharan and Matena, Michael and Zhou, Yanqi and Li, Wei and Liu, Peter J},
  journal={Journal of machine learning research},
  volume={21},
  number={140},
  pages={1--67},
  year={2020}
}

@article{nguyen2023culturax,
  title={Culturax: A cleaned, enormous, and multilingual dataset for large language models in 167 languages},
  author={Nguyen, Thuat and Van Nguyen, Chien and Lai, Viet Dac and Man, Hieu and Ngo, Nghia Trung and Dernoncourt, Franck and Rossi, Ryan A and Nguyen, Thien Huu},
  journal={arXiv preprint arXiv:2309.09400},
  year={2023}
}

@inproceedings{lee-etal-2022-deduplicating,
    title = "Deduplicating Training Data Makes Language Models Better",
    author = "Lee, Katherine  and
      Ippolito, Daphne  and
      Nystrom, Andrew  and
      Zhang, Chiyuan  and
      Eck, Douglas  and
      Callison-Burch, Chris  and
      Carlini, Nicholas",
    editor = "Muresan, Smaranda  and
      Nakov, Preslav  and
      Villavicencio, Aline",
    booktitle = "Proceedings of the 60th Annual Meeting of the Association for Computational Linguistics (Volume 1: Long Papers)",
    month = may,
    year = "2022",
    address = "Dublin, Ireland",
    publisher = "Association for Computational Linguistics",
    url = "https://aclanthology.org/2022.acl-long.577/",
    doi = "10.18653/v1/2022.acl-long.577",
    pages = "8424--8445"
}

@inproceedings{subramani-etal-2023-detecting,
    title = "Detecting Personal Information in Training Corpora: an Analysis",
    author = "Subramani, Nishant  and
      Luccioni, Sasha  and
      Dodge, Jesse  and
      Mitchell, Margaret",
    editor = "Ovalle, Anaelia  and
      Chang, Kai-Wei  and
      Mehrabi, Ninareh  and
      Pruksachatkun, Yada  and
      Galystan, Aram  and
      Dhamala, Jwala  and
      Verma, Apurv  and
      Cao, Trista  and
      Kumar, Anoop  and
      Gupta, Rahul",
    booktitle = "Proceedings of the 3rd Workshop on Trustworthy Natural Language Processing (TrustNLP 2023)",
    month = jul,
    year = "2023",
    address = "Toronto, Canada",
    publisher = "Association for Computational Linguistics",
    url = "https://aclanthology.org/2023.trustnlp-1.18/",
    doi = "10.18653/v1/2023.trustnlp-1.18",
    pages = "208--220"
}

@article{arnett2024toxicity,
  title={{Toxicity of the Commons: Curating Open-Source Pre-Training Data}},
  author={Arnett, Catherine and Jones, Eliot and Yamshchikov, Ivan P. and Langlais, Pierre-Carl},
  journal={arXiv preprint arXiv:2410.22587},
  url={https://arxiv.org/pdf/2410.22587},
  year={2024}
}

@software{together2023redpajama,
  author = {{Together Computer}},
  title = {RedPajama: An Open Source Recipe to Reproduce LLaMA training dataset},
  month = April,
  year = 2023,
  url = {https://github.com/togethercomputer/RedPajama-Data},
  note = {Accessed 30 Jan. 2025}

}

@inproceedings{de-gibert-etal-2024-new-massive,
    title = "A New Massive Multilingual Dataset for High-Performance Language Technologies",
    author = {de Gibert, Ona  and
      Nail, Graeme  and
      Arefyev, Nikolay  and
      Ba{\~n}{\'o}n, Marta  and
      van der Linde, Jelmer  and
      Ji, Shaoxiong  and
      Zaragoza-Bernabeu, Jaume  and
      Aulamo, Mikko  and
      Ram{\'\i}rez-S{\'a}nchez, Gema  and
      Kutuzov, Andrey  and
      Pyysalo, Sampo  and
      Oepen, Stephan  and
      Tiedemann, J{\"o}rg},
    editor = "Calzolari, Nicoletta  and
      Kan, Min-Yen  and
      Hoste, Veronique  and
      Lenci, Alessandro  and
      Sakti, Sakriani  and
      Xue, Nianwen",
    booktitle = "Proceedings of the 2024 Joint International Conference on Computational Linguistics, Language Resources and Evaluation (LREC-COLING 2024)",
    month = may,
    year = "2024",
    address = "Torino, Italia",
    publisher = "ELRA and ICCL",
    url = "https://aclanthology.org/2024.lrec-main.100",
    pages = "1116--1128",
}

@misc{clark2018thinksolvedquestionanswering,
      title={{Think you have Solved Question Answering? Try ARC, the AI2 Reasoning Challenge}}, 
      author={Peter Clark and Isaac Cowhey and Oren Etzioni and Tushar Khot and Ashish Sabharwal and Carissa Schoenick and Oyvind Tafjord},
      year={2018},
      eprint={1803.05457},
      archivePrefix={arXiv},
      primaryClass={cs.AI},
      url={https://arxiv.org/abs/1803.05457}, 
}

@inproceedings{lai-etal-2023-okapi,
    title = "Okapi: Instruction-tuned Large Language Models in Multiple Languages with Reinforcement Learning from Human Feedback",
    author = "Lai, Viet  and
      Nguyen, Chien  and
      Ngo, Nghia  and
      Nguyen, Thuat  and
      Dernoncourt, Franck  and
      Rossi, Ryan  and
      Nguyen, Thien",
    editor = "Feng, Yansong  and
      Lefever, Els",
    booktitle = "Proceedings of the 2023 Conference on Empirical Methods in Natural Language Processing: System Demonstrations",
    month = dec,
    year = "2023",
    address = "Singapore",
    publisher = "Association for Computational Linguistics",
    url = "https://aclanthology.org/2023.emnlp-demo.28/",
    doi = "10.18653/v1/2023.emnlp-demo.28",
    pages = "318--327"
}

@misc{singh2024globalmmluunderstandingaddressing,
      title={Global MMLU: Understanding and Addressing Cultural and Linguistic Biases in Multilingual Evaluation}, 
      author={Shivalika Singh and Angelika Romanou and Clémentine Fourrier and David I. Adelani and Jian Gang Ngui and Daniel Vila-Suero and Peerat Limkonchotiwat and Kelly Marchisio and Wei Qi Leong and Yosephine Susanto and Raymond Ng and Shayne Longpre and Wei-Yin Ko and Madeline Smith and Antoine Bosselut and Alice Oh and Andre F. T. Martins and Leshem Choshen and Daphne Ippolito and Enzo Ferrante and Marzieh Fadaee and Beyza Ermis and Sara Hooker},
      year={2024},
      eprint={2412.03304},
      archivePrefix={arXiv},
      primaryClass={cs.CL},
      url={https://arxiv.org/abs/2412.03304}, 
}

@misc{mikolov2013efficientestimationwordrepresentations,
      title={Efficient Estimation of Word Representations in Vector Space}, 
      author={Tomas Mikolov and Kai Chen and Greg Corrado and Jeffrey Dean},
      year={2013},
      eprint={1301.3781},
      archivePrefix={arXiv},
      primaryClass={cs.CL},
      url={https://arxiv.org/abs/1301.3781}, 
}

@inproceedings{pennington-etal-2014-glove,
    title = "{G}lo{V}e: Global Vectors for Word Representation",
    author = "Pennington, Jeffrey  and
      Socher, Richard  and
      Manning, Christopher",
    editor = "Moschitti, Alessandro  and
      Pang, Bo  and
      Daelemans, Walter",
    booktitle = "Proceedings of the 2014 Conference on Empirical Methods in Natural Language Processing ({EMNLP})",
    month = oct,
    year = "2014",
    address = "Doha, Qatar",
    publisher = "Association for Computational Linguistics",
    url = "https://aclanthology.org/D14-1162/",
    doi = "10.3115/v1/D14-1162",
    pages = "1532--1543"
}

@misc{bojanowski2017enrichingwordvectorssubword,
      title={Enriching Word Vectors with Subword Information}, 
      author={Piotr Bojanowski and Edouard Grave and Armand Joulin and Tomas Mikolov},
      year={2017},
      eprint={1607.04606},
      archivePrefix={arXiv},
      primaryClass={cs.CL},
      url={https://arxiv.org/abs/1607.04606}, 
}

@misc{vaswani2023attentionneed,
      title={Attention Is All You Need}, 
      author={Ashish Vaswani and Noam Shazeer and Niki Parmar and Jakob Uszkoreit and Llion Jones and Aidan N. Gomez and Lukasz Kaiser and Illia Polosukhin},
      year={2023},
      eprint={1706.03762},
      archivePrefix={arXiv},
      primaryClass={cs.CL},
      url={https://arxiv.org/abs/1706.03762}, 
}

@misc{devlin2019bertpretrainingdeepbidirectional,
      title={BERT: Pre-training of Deep Bidirectional Transformers for Language Understanding}, 
      author={Jacob Devlin and Ming-Wei Chang and Kenton Lee and Kristina Toutanova},
      year={2019},
      eprint={1810.04805},
      archivePrefix={arXiv},
      primaryClass={cs.CL},
      url={https://arxiv.org/abs/1810.04805}, 
}

@article{radford2018improving,
  title={Improving language understanding by generative pre-training},
  author={Radford, Alec and Narasimhan, Karthik and Salimans, Tim and Sutskever, Ilya},
  year={2018}
}

@misc{lample2019crosslinguallanguagemodelpretraining,
      title={Cross-lingual Language Model Pretraining}, 
      author={Guillaume Lample and Alexis Conneau},
      year={2019},
      eprint={1901.07291},
      archivePrefix={arXiv},
      primaryClass={cs.CL},
      url={https://arxiv.org/abs/1901.07291}, 
}

@inproceedings{xue-etal-2021-mt5,
    title = "m{T}5: A Massively Multilingual Pre-trained Text-to-Text Transformer",
    author = "Xue, Linting  and
      Constant, Noah  and
      Roberts, Adam  and
      Kale, Mihir  and
      Al-Rfou, Rami  and
      Siddhant, Aditya  and
      Barua, Aditya  and
      Raffel, Colin",
    editor = "Toutanova, Kristina  and
      Rumshisky, Anna  and
      Zettlemoyer, Luke  and
      Hakkani-Tur, Dilek  and
      Beltagy, Iz  and
      Bethard, Steven  and
      Cotterell, Ryan  and
      Chakraborty, Tanmoy  and
      Zhou, Yichao",
    booktitle = "Proceedings of the 2021 Conference of the North American Chapter of the Association for Computational Linguistics: Human Language Technologies",
    month = jun,
    year = "2021",
    address = "Online",
    publisher = "Association for Computational Linguistics",
    url = "https://aclanthology.org/2021.naacl-main.41/",
    doi = "10.18653/v1/2021.naacl-main.41",
    pages = "483--498"
}

@inproceedings{OrtizSuarezSagotRomary2019,
  author    = {Pedro Javier {Ortiz Su{\'a}rez} and Beno{\^i}t Sagot and Laurent Romary},
  title     = {Asynchronous pipelines for processing huge corpora on medium to low resource infrastructures},
  series = {Proceedings of the Workshop on Challenges in the Management of Large Corpora (CMLC-7) 2019. Cardiff, 22nd July 2019},
  editor    = {Piotr Bański and Adrien Barbaresi and Hanno Biber and Evelyn Breiteneder and Simon Clematide and Marc Kupietz and Harald L{\"u}ngen and Caroline Iliadi},
  publisher = {Leibniz-Institut f{\"u}r Deutsche Sprache},
  address   = {Mannheim},
  doi       = {10.14618/ids-pub-9021},
  url       = {http://nbn-resolving.de/urn:nbn:de:bsz:mh39-90215},
  pages     = {9 -- 16},
  year      = {2019},
  language  = {en}
}

@inproceedings{AbadjiOrtizSuarezRomaryetal.2021,
  author    = {Julien Abadji and Pedro Javier Ortiz Su{\'a}rez and Laurent Romary and Beno{\^i}t Sagot},
  title     = {Ungoliant: An optimized pipeline for the generation of a very large-scale multilingual web corpus},
  series = {Proceedings of the Workshop on Challenges in the Management of Large Corpora (CMLC-9) 2021. Limerick, 12 July 2021 (Online-Event)},
  editor    = {Harald L{\"u}ngen and Marc Kupietz and Piotr Bański and Adrien Barbaresi and Simon Clematide and Ines Pisetta},
  publisher = {Leibniz-Institut f{\"u}r Deutsche Sprache},
  address   = {Mannheim},
  doi       = {10.14618/ids-pub-10468},
  url       = {https://nbn-resolving.org/urn:nbn:de:bsz:mh39-104688},
  pages     = {1 -- 9},
  year      = {2021},
  language  = {en}
}

@ARTICLE{2022arXiv220106642A,
       author = {{Abadji}, Julien and {Ortiz Suarez}, Pedro and {Romary}, Laurent and {Sagot}, Beno{\^\i}t},
        title = "{Towards a Cleaner Document-Oriented Multilingual Crawled Corpus}",
      journal = {arXiv e-prints},
         year = 2022,
        month = jan,
          eid = {arXiv:2201.06642},
        pages = {arXiv:2201.06642},
archivePrefix = {arXiv},
       eprint = {2201.06642},
 primaryClass = {cs.CL},
       adsurl = {https://ui.adsabs.harvard.edu/abs/2022arXiv220106642A}
}

@misc{kudugunta2023madlad400multilingualdocumentlevellarge,
      title={{MADLAD-400: A Multilingual And Document-Level Large Audited Dataset}}, 
      author={Sneha Kudugunta and Isaac Caswell and Biao Zhang and Xavier Garcia and Christopher A. Choquette-Choo and Katherine Lee and Derrick Xin and Aditya Kusupati and Romi Stella and Ankur Bapna and Orhan Firat},
      year={2023},
      eprint={2309.04662},
      archivePrefix={arXiv},
      primaryClass={cs.CL},
      url={https://arxiv.org/abs/2309.04662}, 
}

@misc{muennighoff2023scalingdataconstrainedlanguagemodels,
      title={Scaling Data-Constrained Language Models}, 
      author={Niklas Muennighoff and Alexander M. Rush and Boaz Barak and Teven Le Scao and Aleksandra Piktus and Nouamane Tazi and Sampo Pyysalo and Thomas Wolf and Colin Raffel},
      year={2023},
      eprint={2305.16264},
      archivePrefix={arXiv},
      primaryClass={cs.CL},
      url={https://arxiv.org/abs/2305.16264}, 
}

@article{held2025optimizing,
  title={Optimizing Pretraining Data Mixtures with LLM-Estimated Utility},
  author={Held, William and Paranjape, Bhargavi and Koura, Punit Singh and Lewis, Mike and Zhang, Frank and Mihaylov, Todor},
  journal={arXiv preprint arXiv:2501.11747},
  year={2025}
}

@inproceedings{Bandarkar_2024,
   title={{The Belebele Benchmark: a Parallel Reading Comprehension Dataset in 122 Language Variants}},
   url={http://dx.doi.org/10.18653/v1/2024.acl-long.44},
   DOI={10.18653/v1/2024.acl-long.44},
   booktitle={Proceedings of the 62nd Annual Meeting of the Association for Computational Linguistics (Volume 1: Long Papers)},
   publisher={Association for Computational Linguistics},
   author={Bandarkar, Lucas and Liang, Davis and Muller, Benjamin and Artetxe, Mikel and Shukla, Satya Narayan and Husa, Donald and Goyal, Naman and Krishnan, Abhinandan and Zettlemoyer, Luke and Khabsa, Madian},
   year={2024},
   pages={749–775} }

@misc{sen2022mintakacomplexnaturalmultilingual,
      title={{Mintaka: A Complex, Natural, and Multilingual Dataset for End-to-End Question Answering}}, 
      author={Priyanka Sen and Alham Fikri Aji and Amir Saffari},
      year={2022},
      eprint={2210.01613},
      archivePrefix={arXiv},
      primaryClass={cs.CL},
      url={https://arxiv.org/abs/2210.01613}, 
}

@inproceedings{conneau-etal-2018-xnli,
    title = "{XNLI}: Evaluating Cross-lingual Sentence Representations",
    author = "Conneau, Alexis  and
      Rinott, Ruty  and
      Lample, Guillaume  and
      Williams, Adina  and
      Bowman, Samuel  and
      Schwenk, Holger  and
      Stoyanov, Veselin",
    editor = "Riloff, Ellen  and
      Chiang, David  and
      Hockenmaier, Julia  and
      Tsujii, Jun{'}ichi",
    booktitle = "Proceedings of the 2018 Conference on Empirical Methods in Natural Language Processing",
    month = oct # "-" # nov,
    year = "2018",
    address = "Brussels, Belgium",
    publisher = "Association for Computational Linguistics",
    url = "https://aclanthology.org/D18-1269/",
    doi = "10.18653/v1/D18-1269",
    pages = "2475--2485"
}

@misc{zhong2023agievalhumancentricbenchmarkevaluating,
      title={AGIEval: A Human-Centric Benchmark for Evaluating Foundation Models}, 
      author={Wanjun Zhong and Ruixiang Cui and Yiduo Guo and Yaobo Liang and Shuai Lu and Yanlin Wang and Amin Saied and Weizhu Chen and Nan Duan},
      year={2023},
      eprint={2304.06364},
      archivePrefix={arXiv},
      primaryClass={cs.CL},
      url={https://arxiv.org/abs/2304.06364}, 
}

@misc{mozannar2019neuralarabicquestionanswering,
      title={Neural Arabic Question Answering}, 
      author={Hussein Mozannar and Karl El Hajal and Elie Maamary and Hazem Hajj},
      year={2019},
      eprint={1906.05394},
      archivePrefix={arXiv},
      primaryClass={cs.CL},
      url={https://arxiv.org/abs/1906.05394}, 
}

@article{sun-etal-2020-investigating,
    title = "Investigating Prior Knowledge for Challenging {C}hinese Machine Reading Comprehension",
    author = "Sun, Kai  and
      Yu, Dian  and
      Yu, Dong  and
      Cardie, Claire",
    editor = "Johnson, Mark  and
      Roark, Brian  and
      Nenkova, Ani",
    journal = "Transactions of the Association for Computational Linguistics",
    volume = "8",
    year = "2020",
    address = "Cambridge, MA",
    publisher = "MIT Press",
    url = "https://aclanthology.org/2020.tacl-1.10/",
    doi = "10.1162/tacl_a_00305",
    pages = "141--155"
}

@misc{huang2023cevalmultilevelmultidisciplinechinese,
      title={C-Eval: A Multi-Level Multi-Discipline Chinese Evaluation Suite for Foundation Models}, 
      author={Yuzhen Huang and Yuzhuo Bai and Zhihao Zhu and Junlei Zhang and Jinghan Zhang and Tangjun Su and Junteng Liu and Chuancheng Lv and Yikai Zhang and Jiayi Lei and Yao Fu and Maosong Sun and Junxian He},
      year={2023},
      eprint={2305.08322},
      archivePrefix={arXiv},
      primaryClass={cs.CL},
      url={https://arxiv.org/abs/2305.08322}, 
}

@misc{li2024cmmlumeasuringmassivemultitask,
      title={CMMLU: Measuring massive multitask language understanding in Chinese}, 
      author={Haonan Li and Yixuan Zhang and Fajri Koto and Yifei Yang and Hai Zhao and Yeyun Gong and Nan Duan and Timothy Baldwin},
      year={2024},
      eprint={2306.09212},
      archivePrefix={arXiv},
      primaryClass={cs.CL},
      url={https://arxiv.org/abs/2306.09212}, 
}

@inproceedings{Cui_2019,
   title={A Span-Extraction Dataset for Chinese Machine Reading Comprehension},
   url={http://dx.doi.org/10.18653/v1/D19-1600},
   DOI={10.18653/v1/d19-1600},
   booktitle={Proceedings of the 2019 Conference on Empirical Methods in Natural Language Processing and the 9th International Joint Conference on Natural Language Processing (EMNLP-IJCNLP)},
   publisher={Association for Computational Linguistics},
   author={Cui, Yiming and Liu, Ting and Che, Wanxiang and Xiao, Li and Chen, Zhipeng and Ma, Wentao and Wang, Shijin and Hu, Guoping},
   year={2019} }

@inproceedings{talmor-etal-2019-commonsenseqa,
    title = "{C}ommonsense{QA}: A Question Answering Challenge Targeting Commonsense Knowledge",
    author = "Talmor, Alon  and
      Herzig, Jonathan  and
      Lourie, Nicholas  and
      Berant, Jonathan",
    editor = "Burstein, Jill  and
      Doran, Christy  and
      Solorio, Thamar",
    booktitle = "Proceedings of the 2019 Conference of the North {A}merican Chapter of the Association for Computational Linguistics: Human Language Technologies, Volume 1 (Long and Short Papers)",
    month = jun,
    year = "2019",
    address = "Minneapolis, Minnesota",
    publisher = "Association for Computational Linguistics",
    url = "https://aclanthology.org/N19-1421/",
    doi = "10.18653/v1/N19-1421",
    pages = "4149--4158"
}

@misc{hardalov2020examsmultisubjecthighschool,
      title={EXAMS: A Multi-Subject High School Examinations Dataset for Cross-Lingual and Multilingual Question Answering}, 
      author={Momchil Hardalov and Todor Mihaylov and Dimitrina Zlatkova and Yoan Dinkov and Ivan Koychev and Preslav Nakov},
      year={2020},
      eprint={2011.03080},
      archivePrefix={arXiv},
      primaryClass={cs.CL},
      url={https://arxiv.org/abs/2011.03080}, 
}

@misc{dhoffschmidt2020fquadfrenchquestionanswering,
      title={{FQuAD: French Question Answering Dataset}}, 
      author={Martin d'Hoffschmidt and Wacim Belblidia and Tom Brendlé and Quentin Heinrich and Maxime Vidal},
      year={2020},
      eprint={2002.06071},
      archivePrefix={arXiv},
      primaryClass={cs.CL},
      url={https://arxiv.org/abs/2002.06071}, 
}

@misc{zellers2019hellaswagmachinereallyfinish,
      title={HellaSwag: Can a Machine Really Finish Your Sentence?}, 
      author={Rowan Zellers and Ari Holtzman and Yonatan Bisk and Ali Farhadi and Yejin Choi},
      year={2019},
      eprint={1905.07830},
      archivePrefix={arXiv},
      primaryClass={cs.CL},
      url={https://arxiv.org/abs/1905.07830}, 
}

@misc{zhang2023m3exammultilingualmultimodalmultilevel,
      title={M3Exam: A Multilingual, Multimodal, Multilevel Benchmark for Examining Large Language Models}, 
      author={Wenxuan Zhang and Sharifah Mahani Aljunied and Chang Gao and Yew Ken Chia and Lidong Bing},
      year={2023},
      eprint={2306.05179},
      archivePrefix={arXiv},
      primaryClass={cs.CL},
      url={https://arxiv.org/abs/2306.05179}, 
}

@misc{lewis2020mlqaevaluatingcrosslingualextractive,
      title={MLQA: Evaluating Cross-lingual Extractive Question Answering}, 
      author={Patrick Lewis and Barlas Oğuz and Ruty Rinott and Sebastian Riedel and Holger Schwenk},
      year={2020},
      eprint={1910.07475},
      archivePrefix={arXiv},
      primaryClass={cs.CL},
      url={https://arxiv.org/abs/1910.07475}, 
}

@inproceedings{hu-etal-2020-ocnli,
    title = "{OCNLI}: {O}riginal {C}hinese {N}atural {L}anguage {I}nference",
    author = {Hu, Hai  and
      Richardson, Kyle  and
      Xu, Liang  and
      Li, Lu  and
      K{\"u}bler, Sandra  and
      Moss, Lawrence},
    editor = "Cohn, Trevor  and
      He, Yulan  and
      Liu, Yang",
    booktitle = "Findings of the Association for Computational Linguistics: EMNLP 2020",
    month = nov,
    year = "2020",
    address = "Online",
    publisher = "Association for Computational Linguistics",
    url = "https://aclanthology.org/2020.findings-emnlp.314/",
    doi = "10.18653/v1/2020.findings-emnlp.314",
    pages = "3512--3526"
}

@misc{mihaylov2018suitarmorconductelectricity,
      title={Can a Suit of Armor Conduct Electricity? A New Dataset for Open Book Question Answering}, 
      author={Todor Mihaylov and Peter Clark and Tushar Khot and Ashish Sabharwal},
      year={2018},
      eprint={1809.02789},
      archivePrefix={arXiv},
      primaryClass={cs.CL},
      url={https://arxiv.org/abs/1809.02789}, 
}

@misc{bisk2019piqareasoningphysicalcommonsense,
      title={PIQA: Reasoning about Physical Commonsense in Natural Language}, 
      author={Yonatan Bisk and Rowan Zellers and Ronan Le Bras and Jianfeng Gao and Yejin Choi},
      year={2019},
      eprint={1911.11641},
      archivePrefix={arXiv},
      primaryClass={cs.CL},
      url={https://arxiv.org/abs/1911.11641}, 
}

@inproceedings{almazrouei-etal-2023-alghafa,
    title = "{A}l{G}hafa Evaluation Benchmark for {A}rabic Language Models",
    author = "Almazrouei, Ebtesam  and
      Cojocaru, Ruxandra  and
      Baldo, Michele  and
      Malartic, Quentin  and
      Alobeidli, Hamza  and
      Mazzotta, Daniele  and
      Penedo, Guilherme  and
      Campesan, Giulia  and
      Farooq, Mugariya  and
      Alhammadi, Maitha  and
      Launay, Julien  and
      Noune, Badreddine",
    editor = "Sawaf, Hassan  and
      El-Beltagy, Samhaa  and
      Zaghouani, Wajdi  and
      Magdy, Walid  and
      Abdelali, Ahmed  and
      Tomeh, Nadi  and
      Abu Farha, Ibrahim  and
      Habash, Nizar  and
      Khalifa, Salam  and
      Keleg, Amr  and
      Haddad, Hatem  and
      Zitouni, Imed  and
      Mrini, Khalil  and
      Almatham, Rawan",
    booktitle = "Proceedings of ArabicNLP 2023",
    month = dec,
    year = "2023",
    address = "Singapore (Hybrid)",
    publisher = "Association for Computational Linguistics",
    url = "https://aclanthology.org/2023.arabicnlp-1.21/",
    doi = "10.18653/v1/2023.arabicnlp-1.21",
    pages = "244--275"
}

@inproceedings{joshi-etal-2017-triviaqa,
    title = "{T}rivia{QA}: A Large Scale Distantly Supervised Challenge Dataset for Reading Comprehension",
    author = "Joshi, Mandar  and
      Choi, Eunsol  and
      Weld, Daniel  and
      Zettlemoyer, Luke",
    editor = "Barzilay, Regina  and
      Kan, Min-Yen",
    booktitle = "Proceedings of the 55th Annual Meeting of the Association for Computational Linguistics (Volume 1: Long Papers)",
    month = jul,
    year = "2017",
    address = "Vancouver, Canada",
    publisher = "Association for Computational Linguistics",
    url = "https://aclanthology.org/P17-1147/",
    doi = "10.18653/v1/P17-1147",
    pages = "1601--1611"
}

@misc{clark2020tydiqabenchmarkinformationseeking,
      title={{TyDi QA: A Benchmark for Information-Seeking Question Answering in Typologically Diverse Languages}}, 
      author={Jonathan H. Clark and Eunsol Choi and Michael Collins and Dan Garrette and Tom Kwiatkowski and Vitaly Nikolaev and Jennimaria Palomaki},
      year={2020},
      eprint={2003.05002},
      archivePrefix={arXiv},
      primaryClass={cs.CL},
      url={https://arxiv.org/abs/2003.05002}, 
}

@article{sakaguchi2019winogrande,
    title={{WinoGrande: An Adversarial Winograd Schema Challenge at Scale}},
    author={Sakaguchi, Keisuke and Bras, Ronan Le and Bhagavatula, Chandra and Choi, Yejin},
    journal={arXiv preprint arXiv:1907.10641},
    year={2019}
}

@inproceedings{lin-etal-2021-common,
    title = "Common Sense Beyond {E}nglish: Evaluating and Improving Multilingual Language Models for Commonsense Reasoning",
    author = "Lin, Bill Yuchen  and
      Lee, Seyeon  and
      Qiao, Xiaoyang  and
      Ren, Xiang",
    editor = "Zong, Chengqing  and
      Xia, Fei  and
      Li, Wenjie  and
      Navigli, Roberto",
    booktitle = "Proceedings of the 59th Annual Meeting of the Association for Computational Linguistics and the 11th International Joint Conference on Natural Language Processing (Volume 1: Long Papers)",
    month = aug,
    year = "2021",
    address = "Online",
    publisher = "Association for Computational Linguistics",
    url = "https://aclanthology.org/2021.acl-long.102/",
    doi = "10.18653/v1/2021.acl-long.102",
    pages = "1274--1287"
}

@inproceedings{ponti-etal-2020-xcopa,
    title = "{XCOPA}: A Multilingual Dataset for Causal Commonsense Reasoning",
    author = "Ponti, Edoardo Maria  and
      Glava{\v{s}}, Goran  and
      Majewska, Olga  and
      Liu, Qianchu  and
      Vuli{\'c}, Ivan  and
      Korhonen, Anna",
    editor = "Webber, Bonnie  and
      Cohn, Trevor  and
      He, Yulan  and
      Liu, Yang",
    booktitle = "Proceedings of the 2020 Conference on Empirical Methods in Natural Language Processing (EMNLP)",
    month = nov,
    year = "2020",
    address = "Online",
    publisher = "Association for Computational Linguistics",
    url = "https://aclanthology.org/2020.emnlp-main.185/",
    doi = "10.18653/v1/2020.emnlp-main.185",
    pages = "2362--2376"
}

@misc{upadhyay2023xnli20improvingxnli,
      title={XNLI 2.0: Improving XNLI dataset and performance on Cross Lingual Understanding (XLU)}, 
      author={Ankit Kumar Upadhyay and Harsit Kumar Upadhya},
      year={2023},
      eprint={2301.06527},
      archivePrefix={arXiv},
      primaryClass={cs.CL},
      url={https://arxiv.org/abs/2301.06527}, 
}

@article{DBLP:journals/corr/abs-2112-10668,
  author    = {Xi Victoria Lin and
               Todor Mihaylov and
               Mikel Artetxe and
               Tianlu Wang and
               Shuohui Chen and
               Daniel Simig and
               Myle Ott and
               Naman Goyal and
               Shruti Bhosale and
               Jingfei Du and
               Ramakanth Pasunuru and
               Sam Shleifer and
               Punit Singh Koura and
               Vishrav Chaudhary and
               Brian O'Horo and
               Jeff Wang and
               Luke Zettlemoyer and
               Zornitsa Kozareva and
               Mona T. Diab and
               Veselin Stoyanov and
               Xian Li},
  title     = {Few-shot Learning with Multilingual Language Models},
  journal   = {CoRR},
  volume    = {abs/2112.10668},
  year      = {2021},
  url       = {https://arxiv.org/abs/2112.10668},
  eprinttype = {arXiv},
  eprint    = {2112.10668},
  bibsource = {dblp computer science bibliography, https://dblp.org}
}

@misc{tikhonov2021itsheadsusingattention,
      title={It's All in the Heads: Using Attention Heads as a Baseline for Cross-Lingual Transfer in Commonsense Reasoning}, 
      author={Alexey Tikhonov and Max Ryabinin},
      year={2021},
      eprint={2106.12066},
      archivePrefix={arXiv},
      primaryClass={cs.CL},
      url={https://arxiv.org/abs/2106.12066}, 
}

@misc{chinesesquad, 
    title={pluto-junzeng/ChineseSquad}, 
    url={https://github.com/pluto-junzeng/ChineseSquad}, 
    journal={GitHub}, 
    author={Pluto-Junzeng},
    year={2019},
    note = {Accessed 30 Jan. 2025}
}

@misc{koto2024arabicmmluassessingmassivemultitask,
      title={ArabicMMLU: Assessing Massive Multitask Language Understanding in Arabic}, 
      author={Fajri Koto and Haonan Li and Sara Shatnawi and Jad Doughman and Abdelrahman Boda Sadallah and Aisha Alraeesi and Khalid Almubarak and Zaid Alyafeai and Neha Sengupta and Shady Shehata and Nizar Habash and Preslav Nakov and Timothy Baldwin},
      year={2024},
      eprint={2402.12840},
      archivePrefix={arXiv},
      primaryClass={cs.CL},
      url={https://arxiv.org/abs/2402.12840}, 
}

@misc{chang2023multilingualitycurselanguagemodeling,
      title={When Is Multilinguality a Curse? Language Modeling for 250 High- and Low-Resource Languages}, 
      author={Tyler A. Chang and Catherine Arnett and Zhuowen Tu and Benjamin K. Bergen},
      year={2023},
      eprint={2311.09205},
      archivePrefix={arXiv},
      primaryClass={cs.CL},
      url={https://arxiv.org/abs/2311.09205}, 
}

@article{DESAI2024100966,
title = {An archival perspective on pretraining data},
journal = {Patterns},
volume = {5},
number = {4},
pages = {100966},
year = {2024},
issn = {2666-3899},
doi = {https://doi.org/10.1016/j.patter.2024.100966},
url = {https://www.sciencedirect.com/science/article/pii/S2666389924000746},
author = {Meera A. Desai and Irene V. Pasquetto and Abigail Z. Jacobs and Dallas Card}
}

@article{martins2025eurollm,
  title={Eurollm-9b: Technical report},
  author={Martins, Pedro Henrique and Alves, Jo{\~a}o and Fernandes, Patrick and Guerreiro, Nuno M and Rei, Ricardo and Farajian, Amin and Klimaszewski, Mateusz and Alves, Duarte M and Pombal, Jos{\'e} and Boizard, Nicolas and others},
  journal={arXiv preprint arXiv:2506.04079},
  year={2025}
}

@inproceedings{finkelstein2024introducing,
  title={Introducing the NewsPaLM MBR and QE Dataset: LLM-Generated High-Quality Parallel Data Outperforms Traditional Web-Crawled Data},
  author={Finkelstein, Mara and Vilar, David and Freitag, Markus},
  booktitle={Proceedings of the Ninth Conference on Machine Translation},
  pages={1355--1372},
  year={2024}
}

@inproceedings{peter2023there,
  title={There’s No Data like Better Data: Using QE Metrics for MT Data Filtering},
  author={Peter, Jan-Thorsten and Vilar, David and Deutsch, Daniel and Finkelstein, Mara and Juraska, Juraj and Freitag, Markus},
  booktitle={Proceedings of the Eighth Conference on Machine Translation},
  pages={561--577},
  year={2023}
}

\newpage
\section*{NeurIPS Paper Checklist}

\begin{enumerate}

\item {\bf Claims}
    \item[] Question: Do the main claims made in the abstract and introduction accurately reflect the paper's contributions and scope?
    \item[] Answer: \answerYes{} %
    \item[] Justification: Claims are backed by experimental results in Section~\ref{sec:experiments}.

\item {\bf Limitations}
    \item[] Question: Does the paper discuss the limitations of the work performed by the authors?
    \item[] Answer: \answerYes{} %
    \item[] Justification: We discuss the limitations in Appendix~\ref{app:limitations}.

\item {\bf Theory assumptions and proofs}
    \item[] Question: For each theoretical result, does the paper provide the full set of assumptions and a complete (and correct) proof?
    \item[] Answer: \answerNA{} %
    \item[] Justification: The paper does not include theoretical results.

    \item {\bf Experimental result reproducibility}
    \item[] Question: Does the paper fully disclose all the information needed to reproduce the main experimental results of the paper to the extent that it affects the main claims and/or conclusions of the paper (regardless of whether the code and data are provided or not)?
    \item[] Answer: \answerYes{} %
    \item[] Justification: We provide a detailed description of the methods in Section~\ref{sec:method}, experimental setup in Section~\ref{sec:experiments}, evaluation benchmarks selection in Appendix~\ref{app:benchmarks}, and the codebase.

\item {\bf Open access to data and code}
    \item[] Question: Does the paper provide open access to the data and code, with sufficient instructions to faithfully reproduce the main experimental results, as described in supplemental material?
    \item[] Answer: \answerYes{} %
    \item[] Justification: We provide the dataset and the codebase with sufficient information to reproduce the main experimental results.

\item {\bf Experimental setting/details}
    \item[] Question: Does the paper specify all the training and test details (e.g., data splits, hyperparameters, how they were chosen, type of optimizer, etc.) necessary to understand the results?
    \item[] Answer: \answerYes{} %
    \item[] Justification: We provide all details in Section~\ref{sec:method} and Section~\ref{sec:experiments}.

\item {\bf Experiment statistical significance}
    \item[] Question: Does the paper report error bars suitably and correctly defined or other appropriate information about the statistical significance of the experiments?
    \item[] Answer: \answerNo{} %
    \item[] Justification: \citet{penedo2024fineweb-2} show that standard deviation when pretraining large language models in different languages on multiple seeds is small. We do not provide error bars because training large language models is expensive and we use our compute allocation budget for the main experiments. 

\item {\bf Experiments compute resources}
    \item[] Question: For each experiment, does the paper provide sufficient information on the computer resources (type of compute workers, memory, time of execution) needed to reproduce the experiments?
    \item[] Answer: \answerYes{} %
    \item[] Justification: We provide the type of compute workers and the number of GPU hours used in our experiments in Section~\ref{sec:experiments}, and the storage usage of the data in Appendix~\ref{app:sec:dataset}.
    
\item {\bf Code of ethics}
    \item[] Question: Does the research conducted in the paper conform, in every respect, with the NeurIPS Code of Ethics \url{https://neurips.cc/public/EthicsGuidelines}?
    \item[] Answer: \answerYes{} %
    \item[] Justification: We conduct our research in accordance with the NeurIPS Code of Ethics.

\item {\bf Broader impacts}
    \item[] Question: Does the paper discuss both potential positive societal impacts and negative societal impacts of the work performed?
    \item[] Answer: \answerNA{} %
    \item[] Justification: We do not introduce societal impacts beyond those already associated with large language models.
    
\item {\bf Safeguards}
    \item[] Question: Does the paper describe safeguards that have been put in place for responsible release of data or models that have a high risk for misuse (e.g., pretrained language models, image generators, or scraped datasets)?
    \item[] Answer: \answerNo{} %
    \item[] Justification: We base our dataset on the FineWeb-2 dataset which conforms to Common Crawl robots.txt opt-outs (at crawl time), removes personally identifiable content, and offers a form for requesting data removal.

\item {\bf Licenses for existing assets}
    \item[] Question: Are the creators or original owners of assets (e.g., code, data, models), used in the paper, properly credited and are the license and terms of use explicitly mentioned and properly respected?
    \item[] Answer: \answerYes{} %
    \item[] Justification: We credit all assets with their licenses in Appendix~\ref{app:licenses} and respect their licenses.

\item {\bf New assets}
    \item[] Question: Are new assets introduced in the paper well documented and is the documentation provided alongside the assets?
    \item[] Answer: \answerYes{} %
    \item[] Justification: We document the experiments as a part of the dataset creation process in Section~\ref{sec:experiments} and include dataset information in Appendix~\ref{app:sec:dataset}.

\item {\bf Crowdsourcing and research with human subjects}
    \item[] Question: For crowdsourcing experiments and research with human subjects, does the paper include the full text of instructions given to participants and screenshots, if applicable, as well as details about compensation (if any)? 
    \item[] Answer: \answerNA{} %
    \item[] Justification: We do not perform crowdsourcing nor research with human subjects.

\item {\bf Institutional review board (IRB) approvals or equivalent for research with human subjects}
    \item[] Question: Does the paper describe potential risks incurred by study participants, whether such risks were disclosed to the subjects, and whether Institutional Review Board (IRB) approvals (or an equivalent approval/review based on the requirements of your country or institution) were obtained?
    \item[] Answer: \answerNA{} %
    \item[] Justification:  We do not perform research with human subjects.

\item {\bf Declaration of LLM usage}
    \item[] Question: Does the paper describe the usage of LLMs if it is an important, original, or non-standard component of the core methods in this research? Note that if the LLM is used only for writing, editing, or formatting purposes and does not impact the core methodology, scientific rigorousness, or originality of the research, declaration is not required.
    \item[] Answer: \answerNA{} %
    \item[] Justification: Beyond pretraining and evaluating large language models, the core methodology does not involve the use of large language models.

\end{enumerate}

\newpage
\appendix
\section{Dataset Information}\label{app:sec:dataset}
Based on the results of our experiments, we create the dataset, named \textbf{\emph{FineWeb2-HQ}}, by filtering all available FineWeb-2 data (version \texttt{2.0.1}) in 20 languages using the \textit{MLP MKC$^+$} approach with 10\% retention rate. The statistics of the resulting dataset are presented in Table~\ref{tab:dataset_statistics}. We release the dataset under the \textit{Open Data Commons Attribution License (ODC-By) v1.0} license at \href{https://huggingface.co/datasets/epfml/FineWeb2-HQ}{huggingface.co/datasets/epfml/FineWeb2-HQ}.
\\
The main use case of our dataset is LLM pretraining, however, the dataset may also be used for other natural language processing tasks.

\begin{table}[!ht]
    \caption{Statistics (number of documents and disk size) of the dataset resulting from filtering FineWeb-2 using the \textit{MLP MKC$^+$} approach with 10\% retention rate in 20 languages.}
    \centering\scalebox{0.92}{
    \begin{tabular}{lrr}
    \toprule
        Language & Number of documents & Disk size \\
    \midrule
        Russian & 55,220,956 & 1.2TB \\
        Chinese & 54,211,986 & 784GB \\
        German & 43,095,728 & 618GB \\ 
        Spanish & 40,057,637 & 515GB \\
        Japanese & 34,185,427 & 393GB \\
        French & 32,248,772 & 483GB \\
        Italian & 21,180,304 & 269GB \\
        Portuguese & 18,135,468 & 222GB \\
        Polish & 13,384,885 & 168GB \\
        Dutch & 12,920,963 & 160GB \\
        Indonesian & 8,911,149 & 125GB \\
        Turkish & 8,578,808 & 100GB \\
        Czech & 5,995,459 & 104GB \\ 
        Arabic & 5,560,599 & 94GB \\
        Persian & 5,107,187 & 69GB \\
        Hungarian & 4,527,332 & 79GB \\
        Swedish & 4,382,454 & 61GB \\
        Greek & 4,346,440 & 84GB \\
        Danish & 4,082,751 & 61GB \\
        Vietnamese & 4,003,956 & 59GB \\
    \bottomrule
    \end{tabular}}
    \label{tab:dataset_statistics}
\end{table}

\section{Limitations}\label{app:limitations}

A limitation of our work is that we perform experiments on relatively small 1B models with one seed per experiment. We use 1B models to balance the trade-off between the cost of pretraining and the measured signal from the experiments, as found in prior work~\citep{penedo2024fineweb,penedo2024fineweb-2,li2024datacomp}. Additionally, we compare our method to one multilingual baseline, FineWeb-2. However, since FineWeb-2 is the current state-of-the-art and due to our limited computational budget, we decided to allocate more compute towards understanding the mechanics of the data selection process, rather than confirming our results across previous datasets. Nevertheless, computational constraints prevented us from ablating every decision—such as our choice to use only the first 512 tokens for classification. We assume that if the first 512 tokens demonstrate good quality, the remainder of the document likely does as well. Given the strong performance achieved using the first 512 tokens, we prioritized this methodological simplicity. To facilitate further exploration of alternative selection strategies, we have made FineWeb2-embedded\footnote{\href{https://huggingface.co/datasets/epfml/FineWeb2-embedded}{huggingface.co/datasets/epfml/FineWeb2-embedded}\label{footnote:fineweb2-embedded}} available to the community, which contains embeddings for all 512-token chunks.

Although we develop our framework on languages from diverse language families, with different writing systems and with varying resource availability to find an approach that best generalizes for general web crawl text data across languages, classifier training datasets have no quality guarantees for other languages and may result in performance differences that are not visible in our experiments.

Since we focus on simple methods with broad availability and low computational cost, we discuss the computational cost difference between FastText and Transformer embeddings-based methods. While FastText classifiers are cheap to train and inference and can be efficiently run on CPU, Transformer-based methods require an initial computation of embeddings. To mitigate the higher cost of Transformer embeddings, we use a relatively small XLM-RoBERTa model and additionally release the dataset with precomputed embeddings\footref{footnote:fineweb2-embedded}. The total cost for computing the embeddings is approximately 4K compute hours for the 20 languages.

We base our dataset on the FineWeb-2 dataset which conforms to Common Crawl robots.txt opt-outs (at crawl time), removes personally identifiable content, and offers a form for requesting data removal. Since ensuring privacy and fairness of our dataset is beyond the scope of this work, we make the dataset publicly available. This allows other researchers and the public to analyze potential biases, a critical task given that data curation is a political process that can introduce cultural and political impacts~\citep{DESAI2024100966}.

\section{Additional Results}\label{app:additional_results}
\subsection{Model Selection - Per Language Results}\label{app:model_selection}
For clarity, we present the individual benchmark results of the 1B-parameter model trained on 119B tokens for each language in the following tables: Table~\ref{tab:ranking_threshold_10_cmn_Hani_119} for Chinese, Table~\ref{tab:ranking_threshold_10_fra_Latn_119} for French, Table~\ref{tab:ranking_threshold_10_deu_Latn_119} for German, Table~\ref{tab:ranking_threshold_10_arb_Arab_119} for Arabic, and Table~\ref{tab:ranking_threshold_10_dan_Latn_119} for Danish.
    
\begin{table}[!htb]\small
\caption{
Chinese Benchmark performance comparison: Average rank between FineWeb-2 baseline and our proposed filtering methods (\emph{FT}, \emph{MLP}, and \emph{CS}) trained on \emph{MKC$^+$} or \emph{MKC}, retaining top 10\% of documents. The average rank is computed across FineTasks for 1B-parameter models evaluated after 119B tokens.
}
\label{tab:ranking_threshold_10_cmn_Hani_119}
\begin{center}
\begin{tabular}{l>{\columncolor{Best}}lllll>{\columncolor{Baseline}}ll}
\toprule
Approach & \emph{MLP MKC$^+$} & \emph{MLP MKC} & \emph{CS MKC} & \emph{FT MKC} & \emph{FT MKC$^+$} & Baseline & \emph{CS MKC$^+$} \\
\midrule
\rowcolor{AverageRank} Average Rank & 1.7333 & 2.4333 & 4.0667 & 4.0667 & 4.4667 & 5.2333 & 6.0000 \\
AGIEval & \textbf{0.2995} & 0.2948 & 0.2897 & 0.2919 & 0.2817 & 0.2853 & 0.2773 \\
Belebele & \textbf{0.3300} & 0.3233 & 0.3178 & 0.3133 & 0.3133 & 0.3056 & 0.3022 \\
C$^3$ & \textbf{0.4550} & 0.4480 & 0.4400 & 0.4500 & 0.4400 & 0.4400 & 0.4370 \\
C-Eval & \textbf{0.3095} & 0.3060 & 0.2760 & 0.2903 & 0.2906 & 0.2878 & 0.2805 \\
CMMLU & \textbf{0.3312} & 0.3259 & 0.3041 & 0.3043 & 0.3060 & 0.3009 & 0.2995 \\
CMRC 2018 & 0.2224 & 0.2125 & 0.1614 & \textbf{0.2251} & 0.2164 & 0.1949 & 0.1866 \\
HellaSwag & 0.3790 & \textbf{0.3800} & 0.3530 & 0.3680 & 0.3660 & 0.3510 & 0.3370 \\
M3Exam & \textbf{0.3319} & 0.3245 & 0.3084 & 0.3201 & 0.3245 & 0.3216 & 0.3245 \\
X-CODAH & 0.3033 & 0.3000 & \textbf{0.3233} & 0.3100 & 0.2900 & 0.2967 & 0.3067 \\
X-CSQA & \textbf{0.2740} & 0.2680 & 0.2690 & 0.2610 & 0.2520 & 0.2510 & 0.2650 \\
XCOPA & 0.6200 & \textbf{0.6400} & 0.6180 & 0.5740 & 0.5740 & 0.6000 & 0.5620 \\
OCNLI & 0.5470 & 0.5470 & 0.5340 & 0.5250 & \textbf{0.5600} & 0.5420 & 0.5060 \\
Chinese-SQuAD & 0.0929 & \textbf{0.1097} & 0.0865 & 0.0889 & 0.0850 & 0.0777 & 0.0585 \\
XStoryCloze & \textbf{0.5800} & 0.5630 & 0.5710 & 0.5560 & 0.5610 & 0.5580 & 0.5570 \\
XWINO & 0.6429 & 0.6528 & \textbf{0.6587} & 0.6131 & 0.5992 & 0.6429 & 0.6111 \\
\bottomrule
\end{tabular}
\end{center}
\end{table}

\begin{table}[!htb]\small
\caption{
French Benchmark performance comparison: Average rank between FineWeb-2 baseline and our proposed filtering methods (\emph{FT}, \emph{MLP}, and \emph{CS}) trained on \emph{MKC$^+$} or \emph{MKC}, retaining top 10\% of documents. The average rank is computed across FineTasks for 1B-parameter models evaluated after 119B tokens.
}
\label{tab:ranking_threshold_10_fra_Latn_119}
\begin{center}
\begin{tabular}{ll>{\columncolor{Best}}lllll>{\columncolor{Baseline}}l}
\toprule
Approach & \emph{FT MKC$^+$} & \emph{MLP MKC$^+$} & \emph{MLP MKC} & \emph{FT MKC} & \emph{CS MKC} & \emph{CS MKC$^+$} & Baseline \\
\midrule
\rowcolor{AverageRank} Average Rank & 3.2222 & 3.5000 & 3.5556 & 3.7778 & 4.0000 & 4.6667 & 5.2778 \\
Belebele & 0.3378 & 0.3533 & \textbf{0.3678} & 0.3489 & 0.3444 & 0.3344 & 0.3444 \\
HellaSwag & \textbf{0.5380} & \textbf{0.5380} & 0.4990 & 0.5150 & 0.5280 & 0.5070 & 0.5180 \\
X-CSQA & 0.2820 & 0.2740 & 0.2730 & \textbf{0.2990} & 0.2850 & 0.2900 & 0.2870 \\
XNLI 2.0 & 0.7340 & 0.7400 & 0.7430 & 0.7230 & \textbf{0.7450} & 0.7330 & 0.7180 \\
FQuAD & 0.2597 & 0.2803 & \textbf{0.3032} & 0.2981 & 0.2411 & 0.2476 & 0.2401 \\
MMLU & 0.2896 & 0.2895 & \textbf{0.2925} & 0.2886 & 0.2806 & 0.2815 & 0.2706 \\
Mintaka & 0.0710 & 0.0438 & 0.0334 & 0.0670 & 0.0610 & \textbf{0.0976} & 0.0712 \\
X-CODAH & \textbf{0.3000} & 0.2667 & 0.2867 & 0.2767 & \textbf{0.3000} & 0.2800 & 0.2633 \\
ARC (Challenge) & 0.3120 & \textbf{0.3180} & 0.3090 & 0.3060 & 0.2950 & 0.2830 & 0.2850 \\
\bottomrule
\end{tabular}
\end{center}
\end{table}

\begin{table}[!htb]\small

\caption{
German Benchmark performance comparison: Average rank between FineWeb-2 baseline and our proposed filtering methods (\emph{FT}, \emph{MLP}, and \emph{CS}) trained on \emph{MKC$^+$} or \emph{MKC}, retaining top 10\% of documents. The average rank is computed across FineTasks for 1B-parameter models evaluated after 119B tokens.
}
\label{tab:ranking_threshold_10_deu_Latn_119}
\begin{center}
\begin{tabular}{l>{\columncolor{Best}}llllll>{\columncolor{Baseline}}l}
\toprule
Approach & \emph{MLP MKC$^+$} & \emph{FT MKC$^+$} & \emph{FT MKC} & \emph{CS MKC} & \emph{MLP MKC} & \emph{CS MKC$^+$} & Baseline \\
\midrule
\rowcolor{AverageRank} Average Rank & 3.1250 & 3.1250 & 3.5000 & 3.7500 & 4.5000 & 4.7500 & 5.2500 \\
MMLU & \textbf{0.2940} & 0.2879 & 0.2926 & 0.2770 & 0.2905 & 0.2764 & 0.2718 \\
ARC (Challenge) & 0.2760 & 0.2850 & 0.2820 & \textbf{0.2880} & 0.2830 & 0.2640 & 0.2680 \\
Mintaka & 0.0580 & 0.0548 & 0.0735 & 0.0576 & 0.0494 & \textbf{0.0766} & 0.0498 \\
Belebele & \textbf{0.3611} & 0.3578 & 0.3544 & 0.3544 & 0.3567 & 0.3422 & 0.3544 \\
X-CODAH & 0.3367 & 0.3500 & 0.3300 & 0.3567 & 0.3400 & \textbf{0.3600} & 0.3467 \\
X-CSQA & 0.2978 & \textbf{0.3008} & 0.2877 & 0.2887 & 0.2857 & 0.2918 & 0.2787 \\
HellaSwag & 0.4640 & 0.4710 & \textbf{0.4870} & 0.4820 & 0.4540 & 0.4390 & 0.4470 \\
XNLI 2.0 & 0.6620 & 0.6530 & 0.6740 & 0.6440 & 0.6610 & 0.6520 & \textbf{0.6890} \\
\bottomrule
\end{tabular}
\end{center}
\end{table}

\begin{table}[!htb]\small
\caption{
Arabic Benchmark performance comparison: Average rank between FineWeb-2 baseline and our proposed filtering methods (\emph{FT}, \emph{MLP}, and \emph{CS}) trained on \emph{MKC$^+$} or \emph{MKC}, retaining top 56\% of documents. The average rank is computed across FineTasks for 1B-parameter models evaluated after 119B tokens.
}
\label{tab:ranking_threshold_10_arb_Arab_119}
\begin{center}
\begin{tabular}{l>{\columncolor{Best}}lll>{\columncolor{Baseline}}llll}
\toprule
Approach & \emph{MLP MKC$^+$} & \emph{MLP MKC} & \emph{FT MKC$^+$} & Baseline & \emph{CS MKC$^+$} & \emph{CS MKC} & \emph{FT MKC} \\
\midrule
\rowcolor{AverageRank} Average Rank & 2.7812 & 3.2500 & 3.6875 & 3.9688 & 3.9688 & 5.0312 & 5.3125 \\
EXAMS & 0.3537 & \textbf{0.3656} & 0.3552 & 0.3582 & 0.3443 & 0.3262 & 0.3346 \\
MMLU & 0.4007 & 0.3909 & \textbf{0.4023} & 0.3894 & 0.3912 & 0.3781 & 0.3885 \\
ARC (Easy) & \textbf{0.4330} & 0.4230 & 0.4210 & 0.4120 & 0.4020 & 0.3940 & 0.4080 \\
AlGhafa SciQ & 0.6915 & \textbf{0.7005} & 0.6965 & 0.6854 & 0.6724 & 0.6683 & 0.6804 \\
Belebele & 0.3456 & 0.3356 & 0.3322 & 0.3311 & 0.3356 & \textbf{0.3567} & 0.3233 \\
SOQAL & \textbf{0.7333} & 0.6867 & 0.7000 & 0.7200 & 0.7267 & 0.6867 & 0.7133 \\
MLQA & 0.2386 & \textbf{0.2402} & 0.1928 & 0.1901 & 0.2189 & 0.2154 & 0.1793 \\
TyDi QA & \textbf{0.1547} & 0.1476 & 0.1230 & 0.1441 & 0.1223 & 0.1097 & 0.1182 \\
AlGhafa RACE & 0.3720 & \textbf{0.3740} & 0.3640 & 0.3710 & 0.3590 & 0.3660 & 0.3730 \\
ARCD & \textbf{0.3638} & 0.3505 & 0.3235 & 0.3354 & 0.3358 & 0.3432 & 0.3043 \\
X-CODAH & 0.2600 & 0.2533 & 0.2567 & \textbf{0.2633} & \textbf{0.2633} & 0.2500 & 0.2600 \\
AlGhafa PIQA & 0.6360 & 0.6320 & \textbf{0.6400} & 0.6240 & 0.6320 & 0.6320 & 0.6370 \\
X-CSQA & 0.2740 & 0.2810 & 0.2770 & \textbf{0.2900} & 0.2880 & 0.2720 & 0.2770 \\
XNLI 2.0 & 0.6570 & 0.6910 & 0.6990 & \textbf{0.7010} & 0.6910 & 0.6900 & 0.6770 \\
HellaSwag & 0.4270 & 0.4220 & 0.4280 & 0.4250 & 0.4260 & \textbf{0.4320} & 0.4150 \\
XStoryCloze & 0.6150 & 0.6100 & 0.6100 & 0.6070 & 0.6130 & \textbf{0.6180} & 0.5930 \\
\bottomrule
\end{tabular}
\end{center}
\end{table}

\begin{table}[!htb]\small
\caption{
Danish Benchmark performance comparison: Average rank between FineWeb-2 baseline and our proposed filtering methods (\emph{FT}, \emph{MLP}, and \emph{CS}) trained on \emph{MKC$^+$} or \emph{MKC}, retaining top 65\% of documents. The average rank is computed across FineTasks for 1B-parameter models evaluated after 119B tokens.
}
\label{tab:ranking_threshold_10_dan_Latn_119}
\begin{center}
\begin{tabular}{ll>{\columncolor{Best}}ll>{\columncolor{Baseline}}l}
\toprule
Approach & \emph{CS MKC$^+$} & \emph{MLP MKC$^+$} & \emph{FT MKC$^+$} & Baseline \\
\midrule
\rowcolor{AverageRank} Average Rank & 1.0000 & 2.5000 & 3.1667 & 3.3333 \\
ARC (Challenge) & \textbf{0.2820} & 0.2650 & 0.2730 & 0.2560 \\
HellaSwag & \textbf{0.4950} & 0.4850 & 0.4750 & 0.4750 \\
Belebele & \textbf{0.3333} & 0.3289 & 0.3189 & 0.3289 \\
\bottomrule
\end{tabular}
\end{center}
\end{table}

\subsection{Threshold Selection}\label{app:threshold_selection}
\textbf{Complete Result.} To confirm that the \emph{CS} filtering method is not competitive with \emph{MLP} and \emph{FT}, even when a higher percentage of documents is retained, we present the complete threshold selection results, including the \emph{CS} method, in Table~\ref{tab:ranking_threshold_15_20_all} in addition to the results shown in Section~\ref{sec:threshold_selection} (Table~\ref{tab:ranking_threshold_10_15_20}).

\begin{figure}[!htb]
    \vspace{-1em}
    \centering
\includegraphics[width=0.5\linewidth]{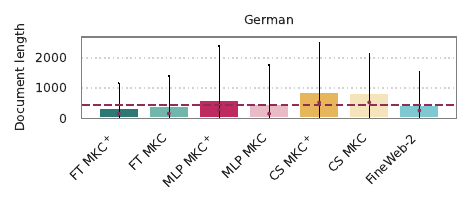}
\caption{
Document length comparison: Average length and standard deviation in FineWeb-2 before and after 10\% retention filtering. Red horizontal line shows average document length, red dots indicate medians. Length measured by space-separated tokens.
}
    \label{fig:doc_length_de}
    \vspace{-1em}
\end{figure}

\begin{figure}[!htb]
    \centering    
    \includegraphics[width=0.97\textwidth]{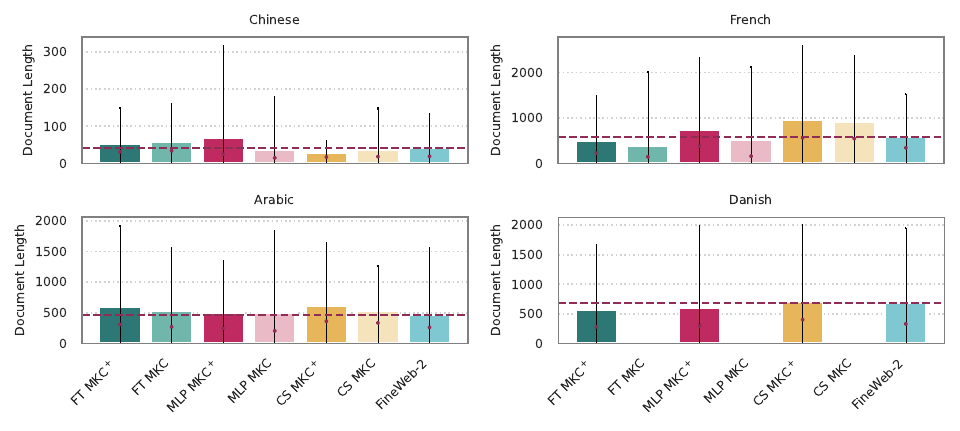}
    \vspace{-1em}
    \caption{Document length comparison: Average length and standard deviation in FineWeb-2 before and after 10\% retention filtering. Red horizontal line shows average document length, red dots indicate medians. Length measured by space-separated tokens.}
    \label{fig:doc_length_other}
\end{figure}

\textbf{Document Length Bias.} Motivated by the observed bias in certain approaches favoring the selection of shorter documents, as seen in Figure~\ref{fig:doc_length_de}, Figure~\ref{fig:doc_length_other} and Table~\ref{tab:token_count_threshold_10_56_65}, we examine how this bias interacts with performance when retaining more documents. As demonstrated in Table~\ref{tab:token_count_threshold_10_56_65}, the \emph{MLP MKC} approach shows a tendency to retain shorter documents, while achieving higher performance with an increased number of retained documents. In contrast, the \emph{CS} and \emph{FT} filtering methods present mixed results, suggesting that the optimal threshold selection may be influenced by additional factors.

\begin{table}[!htb]\small
\caption{
Benchmark performance comparison: Average rank between FineWeb-2 baseline and our proposed filtering methods (\emph{FT}, \emph{MLP}) trained on \emph{MKC$^+$} or \emph{MKC}, retaining top 10\%, 15\% or 20\% of documents. The average rank is computed across FineTasks for 1B-parameter models evaluated on Chinese, German and French after 70B and 119B tokens.    
}
\label{tab:ranking_threshold_15_20_all}
\begin{center}
\begin{tabular}{lcr}
\toprule
Approach & Threshold & Average Rank \\
\midrule
\rowcolor{Best} \emph{MLP MKC$^+$} & 10\% & 11.73 \\
\emph{MLP MKC$^+$} & 15\% & 12.13 \\
\emph{MLP MKC} & 20\% & 15.07 \\
\emph{MLP MKC} & 15\% & 15.09 \\
\emph{MLP MKC$^+$} & 20\% & 15.40 \\
\emph{MLP MKC} & 10\% & 16.09 \\
\emph{FT MKC$^+$} & 10\% & 18.61 \\
\emph{CS MKC} & 15\% & 19.02 \\
\emph{CS MKC} & 20\% & 19.24 \\
\emph{FT MKC} & 15\% & 19.84 \\
\emph{FT MKC} & 10\% & 20.02 \\
\emph{CS MKC} & 10\% & 20.67 \\
\emph{FT MKC} & 20\% & 20.80 \\
\emph{FT MKC$^+$} & 15\% & 22.05 \\
\emph{FT MKC$^+$} & 20\% & 22.52 \\
\emph{CS MKC$^+$} & 15\% & 24.66 \\
\emph{CS MKC$^+$} & 20\% & 25.08 \\
\rowcolor{Baseline} Baseline & -- & 25.54 \\
\emph{CS MKC$^+$} & 10\% & 26.94 \\
\bottomrule
\end{tabular}
\end{center}
\end{table}


\begin{table}[!htb]\small
\caption{
Token retention comparison: Counts in FineWeb-2 before and after filtering using our approach with 10\% document retention for Chinese, French and German, 56\% for Arabic, and 65\% for Danish. Token counts represent tokenized dataset sizes using the multilingual Mistral v3 (Tekken) tokenizer~\citep{tekkenV3}.
}
\label{tab:token_count_threshold_10_56_65}
\begin{center}
\begin{tabular}{lllllr}
\toprule
Approach & Chinese & French & German & Arabic & Danish \\
\midrule
\rowcolor{Best} \emph{MLP MKC$^+$} & 150B (9\%) & 89B (12\%) & 119B (12\%) & 78B (61\%) & 71B (66\%) \\
\emph{MLP MKC} & 105B (7\%) & 72B (10\%) & 87B (9\%) & 75B (59\%) & -- \\
\midrule
\emph{FT MKC$^+$} & 221B (14\%) & 70B (10\%) & 63B (6\% )& 77B (61\%) & 70B (65\%) \\
\emph{FT MKC} & 190B (12\%) & 43B (6\%) & 65B (7\%) & 80B (63\%) & -- \\
\midrule
\emph{CS MKC$^+$} & 170B (11\%) & 126B (17\%) & 166B (17\%) & 82B (65\%) & 77B (71\%) \\
\emph{CS MKC} & 161B (10\%) & 132B (18\%) & 172B (18\%) & 83B (65\%) & -- \\
\midrule
\rowcolor{Baseline} Baseline & 1597B & 730B & 973B & 127B & 108B \\
\bottomrule
\end{tabular}
\end{center}
\end{table}

\subsection{Training Data Analysis}\label{app:training_data_analysis}
\begin{figure}[!htb]
    \centering    
    \includegraphics[width=0.55\textwidth]{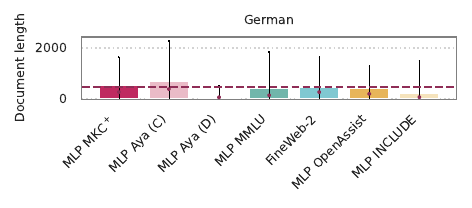}
    \caption{Document length comparison: Average length and standard deviation in FineWeb-2 before and after filtering using \emph{MLP} method with 10\% retention on different training datasets. Red horizontal line shows average document length, red dots indicate medians. Length measured by space-separated tokens.}
    \label{fig:doc_length_other_per_language_german}
\end{figure}

\begin{figure}[!htb]
    \centering    
    \includegraphics[width=0.97\textwidth]{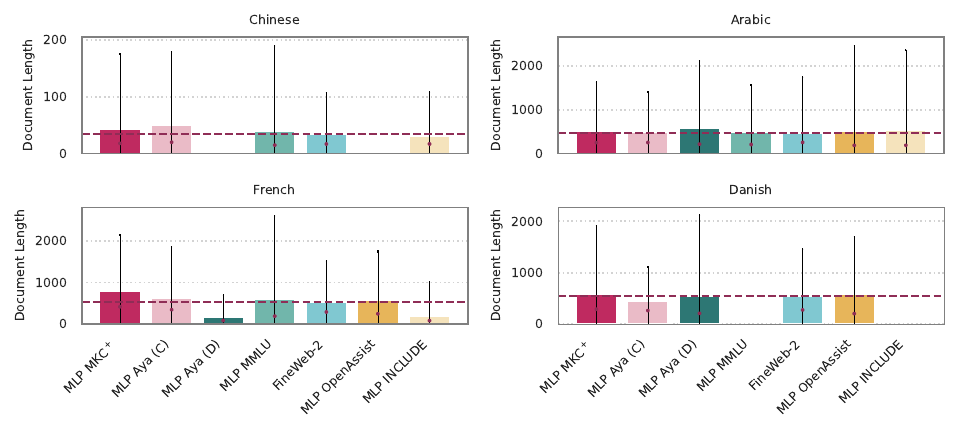}
    \caption{Document length comparison: Average length and standard deviation in FineWeb-2 before and after filtering using \emph{MLP} method with 10\% retention for Chinese and French, 56\% for Arabic and 65\% for Danish on different training datasets. Red horizontal line shows average document length, red dots indicate medians. Length measured by space-separated tokens.}
    \label{fig:doc_length_other_per_language_others}
\end{figure}

We give details on the variation in the average length of documents retained by our model-based filtering method \emph{MLP} for Chinese, French, Arabic, and Danish with different training datasets. The results are shown for German in Figure ~\ref{fig:doc_length_other_per_language_german} and for all other languages in Figure~\ref{fig:doc_length_other_per_language_others}.

\subsection{Replay of Original Data}\label{app:sec:replay}

We explore whether incorporating a small percentage of original raw data (replay) can help improve performance. We do this for our best FastText (\emph{FT MKC$^+$}) and Transformer approaches (\emph{MLP MKC$^+$}). Table \ref{tab:ranking_mixture} presents the results of experiments where 5\% and 10\% unfiltered data were mixed into the training dataset, alongside results from training without any replay. Although, the \emph{FT MKC$^+$} filters shows mixed signal, our \emph{MLP MKC$^+$} approach clearly demonstrates that replay does not improve performance, indicating the data selection already retains enough diversity. In cases of less diverse datasets, replay was shown to offer benefits~\citep{bethune2025scalinglawsforgettingfinetuning,chen2023meditron70bscalingmedicalpretraining}.

\begin{table}[!htb]\small
        \centering
        \caption{
        Benchmark performance comparison: Average rank of our \emph{MLP MKC$^+$} and \emph{FT MKC$^+$} approaches with 10\% document retention, mixed with 0\%, 5\%, or 10\% of original FineWeb-2 dataset. The average rank is computed across FineTasks for 1B-parameter models evaluated on Chinese, German and French after 70B and 119B tokens.
        }
        \label{tab:ranking_mixture}
        \begin{center}
        \begin{tabular}{lrr}
        \toprule
        Approach & Mixture Rate & Average Rank \\
        \midrule
        \emph{MLP MKC$^+$} & 5\% & 5.09 \\
        \rowcolor{Best} \emph{MLP MKC$^+$} & 0\% & 5.16 \\
        \emph{MLP MKC$^+$} & 10\% & 5.40 \\
        \emph{FT MKC$^+$} & 10\% & 7.17 \\
        \emph{FT MKC$^+$} & 0\% & 7.51 \\
        \emph{FT MKC$^+$} & 5\% & 8.66 \\
        \bottomrule
        \end{tabular}
        \end{center}
\end{table}

\subsection{Impact on multilingual model training}\label{app:multilingual}
This section presents the results of our \emph{MLP MKC$^+$} approach on multilingual model training for Chinese (Table~\ref{tab:ranking_multilingual_cmn_Hani_595}), Arabic (Table~\ref{tab:ranking_multilingual_arb_Arab_595}), German (Table~\ref{tab:ranking_multilingual_deu_Latn_595}), and Danish (Table~\ref{tab:ranking_multilingual_dan_Latn_595}), in addition to the results for French discussed in Section~\ref{sec:multilingual_llm}.

\begin{table}[!htb]\small
    \caption{Chinese benchmark performance: Multilingual LLMs ($M$) trained on FineWeb-2 or our \emph{MLP MKC$^+$} refined dataset (retaining top 10\% for Chinese, German and French, 56\% for Arabic, 65\% for Danish) with 595B tokens, compared to monolingual models trained on 119B tokens. The average rank is computed across FineTasks for 1B-parameter models.
    }
    \label{tab:ranking_multilingual_cmn_Hani_595}
    \begin{center}
    \begin{tabular}{l>{\columncolor{Best}}lll>{\columncolor{Baseline}}r}
    \toprule
    Dataset & Ours & Ours$_M$ & FW-2$_M$ & FW-2 \\
    \midrule
    \rowcolor{AverageRank} Average Rank & 1.5667 & 2.1667 & 2.9000 & 3.3667 \\
    AGIEval & \textbf{0.2995} & 0.2863 & 0.2894 & 0.2853 \\
    Belebele & 0.3300 & \textbf{0.3456} & 0.3189 & 0.3056 \\
    C$^3$ & \textbf{0.4550} & 0.4520 & 0.4480 & 0.4400 \\
    C-Eval & \textbf{0.3095} & 0.2848 & 0.2683 & 0.2878 \\
    CMMLU & \textbf{0.3312} & 0.3064 & 0.2967 & 0.3009 \\
    CMRC 2018  & 0.2224 & \textbf{0.2689} & 0.2090 & 0.1949 \\
    HellaSwag & \textbf{0.3790} & 0.3740 & 0.3740 & 0.3510 \\
    M3Exam & \textbf{0.3319} & 0.3040 & 0.3304 & 0.3216 \\
    X-CODAH & 0.3033 & \textbf{0.3067} & 0.2800 & 0.2967 \\
    X-CSQA & 0.2740 & \textbf{0.2810} & 0.2780 & 0.2510 \\
    XCOPA & \textbf{0.6200} & 0.6020 & 0.5860 & 0.6000 \\
    OCNLI & \textbf{0.5470} & 0.5320 & 0.4910 & 0.5420 \\
    Chinese-SQuAD  & 0.0929 & \textbf{0.1304} & 0.1017 & 0.0777 \\
    XStoryCloze & \textbf{0.5800} & 0.5760 & 0.5650 & 0.5580 \\
    XWINO & 0.6429 & 0.6409 & \textbf{0.6468} & 0.6429 \\
    \bottomrule
    \end{tabular}
    \end{center}
\end{table}

\begin{table}[!htb]\small
    \caption{Arabic benchmark performance: Multilingual LLMs ($M$) trained on FineWeb-2 or our \emph{MLP MKC$^+$} refined dataset (retaining top 10\% for Chinese, German and French, 56\% for Arabic, 65\% for Danish) with 595B tokens, compared to monolingual models trained on 119B tokens. The average rank is computed across FineTasks for 1B-parameter models.}
    \label{tab:ranking_multilingual_arb_Arab_595}
    \begin{center}
    \begin{tabular}{ll>{\columncolor{Best}}l>{\columncolor{Baseline}}lr}
    \toprule
    Dataset & Ours$_M$ & Ours & FW-2 & FW-2$_M$ \\
    \midrule
    \rowcolor{AverageRank} Average Rank & 1.9688 & 2.0000 & 2.7500 & 3.2812 \\
    EXAMS & 0.3336 & 0.3537 & \textbf{0.3582} & 0.3076 \\
    MMLU & 0.3828 & \textbf{0.4007} & 0.3894 & 0.3599 \\
    ARC (Easy) & 0.4190 & \textbf{0.4330} & 0.4120 & 0.3760 \\
    AlGhafa SciQ & 0.6764 & \textbf{0.6915} & 0.6854 & 0.6563 \\
    Belebele & \textbf{0.3511} & 0.3456 & 0.3311 & 0.3344 \\
    SOQAL & 0.7000 & \textbf{0.7333} & 0.7200 & 0.6533 \\
    MLQA  & 0.2208 & \textbf{0.2386} & 0.1901 & 0.2085 \\
    TyDi QA  & \textbf{0.1634} & 0.1547 & 0.1441 & 0.1429 \\
    AlGhafa RACE & \textbf{0.3830} & 0.3720 & 0.3710 & 0.3770 \\
    ARCD & 0.3377 & \textbf{0.3638} & 0.3354 & 0.2970 \\
    X-CODAH & \textbf{0.2767} & 0.2600 & 0.2633 & \textbf{0.2767} \\
    AlGhafa PIQA & 0.6170 & \textbf{0.6360} & 0.6240 & 0.6160 \\
    X-CSQA & 0.2860 & 0.2740 & \textbf{0.2900} & 0.2660 \\
    XNLI 2.0 & 0.7080 & 0.6570 & 0.7010 & \textbf{0.7340} \\
    HellaSwag & \textbf{0.4390} & 0.4270 & 0.4250 & 0.4240 \\
    XStoryCloze & \textbf{0.6370} & 0.6150 & 0.6070 & 0.6160 \\
    \bottomrule
    \end{tabular}
    \end{center}
\end{table}

\begin{table}[!htb]\small
    \caption{German benchmark performance: Multilingual LLMs ($M$) trained on FineWeb-2 or our \emph{MLP MKC$^+$} refined dataset (retaining top 10\% for Chinese, German and French, 56\% for Arabic, 65\% for Danish) with 595B tokens, compared to monolingual models trained on 119B tokens. The average rank is computed across FineTasks for 1B-parameter models.}
    \label{tab:ranking_multilingual_deu_Latn_595}
    \begin{center}
    \begin{tabular}{ll>{\columncolor{Best}}l>{\columncolor{Baseline}}lr}
    \toprule
    Dataset & Ours$_M$ & Ours & FW-2 & FW-2$_M$ \\
    \midrule
    \rowcolor{AverageRank} Average Rank & 1.5000 & 2.1250 & 2.9375 & 3.4375 \\
    MMLU & 0.2918 & \textbf{0.2940} & 0.2718 & 0.2691 \\
    ARC (Challenge) & 0.2740 & \textbf{0.2760} & 0.2680 & 0.2640 \\
    Mintaka  & \textbf{0.0821} & 0.0580 & 0.0498 & 0.0660 \\
    Belebele & \textbf{0.3956} & 0.3611 & 0.3544 & 0.3633 \\
    X-CODAH & \textbf{0.3500} & 0.3367 & 0.3467 & 0.3167 \\
    X-CSQA & \textbf{0.3048} & 0.2978 & 0.2787 & 0.2787 \\
    HellaSwag & \textbf{0.4690} & 0.4640 & 0.4470 & 0.4430 \\
    XNLI 2.0 & 0.6420 & 0.6620 & \textbf{0.6890} & 0.6340 \\
    \bottomrule
    \end{tabular}
    \end{center}
\end{table}

\begin{table}[!htb]\small
    \caption{Danish benchmark performance: Multilingual LLMs ($M$) trained on FineWeb-2 or our \emph{MLP MKC$^+$} refined dataset (retaining top 10\% for Chinese, German and French, 56\% for Arabic, 65\% for Danish) with 595B tokens, compared to monolingual models trained on 119B tokens. The average rank is computed across FineTasks for 1B-parameter models.}
    \label{tab:ranking_multilingual_dan_Latn_595}
    \begin{center}
    \begin{tabular}{ll>{\columncolor{Best}}ll>{\columncolor{Baseline}}r}
    \toprule
    Dataset & Ours$_M$ & Ours & FW-2$_M$ & FW-2 \\
    \midrule
    \rowcolor{AverageRank} Average Rank & 1.6667 & 2.1667 & 3.0000 & 3.1667 \\
    ARC (Challenge) & \textbf{0.2920} & 0.2650 & 0.2600 & 0.2560 \\
    HellaSwag & 0.4710 & \textbf{0.4850} & 0.4560 & 0.4750 \\
    Belebele & \textbf{0.3700} & 0.3289 & 0.3311 & 0.3289 \\
    \bottomrule
    \end{tabular}
    \end{center}
\end{table}

\clearpage

\clearpage

\section{List of evaluation benchmarks and metrics}\label{app:benchmarks}
We provide a detailed overview of the evaluation benchmarks used to assess our models' performance, along with their respective evaluation metrics in Table~\ref{tab:benchmark_overview}. For non-English tasks and English MMLU, we use the \emph{cloze} multiple-choice prompt, which allows the model to directly predict each option instead of using the standard prompt format with A/B/C/D letter prefixes as targets. This approach was chosen because it has been shown to serve as a more reliable performance indicator earlier in training~\citep{kydlicek2024finetasksmultilingualtasks}. We evaluate the models in a 0-shot setting.

\begin{table}[!htb]\small
    \caption{List of Evaluation Benchmarks and Metrics used in our setup for Chinese, French, German, Arabic, Danish, and English.}\label{tab:benchmark_overview}
    \begin{center}
\resizebox{0.99\textwidth}{!}{
    \begin{tabular}{|l|c|c|c|c|c|c|c|}
    \hline
        Benchmark & Chinese & French & German & Arabic & Danish & English & Evaluation metric \\ \hline
        AGIEval~\citep{zhong2023agievalhumancentricbenchmarkevaluating} & \checkmark &  &  &  &  &  & Normalized accuracy \\ \hline
        AlGhafa ARC~\citep{almazrouei-etal-2023-alghafa} &  & & & \checkmark & &  & Normalized accuracy \\ \hline
        AlGhafa PIQA~\citep{almazrouei-etal-2023-alghafa} &  &  &  & \checkmark &  &  & Normalized accuracy \\ \hline
        AlGhafa RACE~\citep{almazrouei-etal-2023-alghafa} &  &  &  & \checkmark &  &  & Normalized accuracy \\ \hline
        AlGhafa SciQ~\citep{almazrouei-etal-2023-alghafa} &  &  &  & \checkmark &  &  & Normalized accuracy \\ \hline    ArabicMMLU~\citep{koto2024arabicmmluassessingmassivemultitask} &  &  &  & \checkmark &  &  & Normalized accuracy \\ \hline
        ARC~\citep{clark2018thinksolvedquestionanswering} &  &  &  &  &  & \checkmark & Normalized accuracy \\ \hline
        ARCD~\citep{mozannar2019neuralarabicquestionanswering} &  &  &  & \checkmark &  &  & F1 score \\ \hline
        Belebele~\citep{Bandarkar_2024} & \checkmark & \checkmark & \checkmark & \checkmark & \checkmark &  & Normalized accuracy \\ \hline
        C$^3$~\citep{sun-etal-2020-investigating} & \checkmark &  &  &  &  &  & Normalized accuracy \\ \hline
        C-Eval~\citep{huang2023cevalmultilevelmultidisciplinechinese} & \checkmark &  &  &  &  &  & Normalized accuracy \\ \hline
        Chinese-SQuAD~\citep{chinesesquad} & \checkmark &  &  &  &  &  & F1 score \\ \hline
        CMMLU~\citep{li2024cmmlumeasuringmassivemultitask} & \checkmark &  &  &  &  &  & Normalized accuracy \\ \hline
        CMRC 2018~\citep{Cui_2019} & \checkmark &  &  &  &  &  & F1 score \\ \hline
        CommonsenseQA~\citep{talmor-etal-2019-commonsenseqa} &  &  &  &  &  & \checkmark & Normalized accuracy \\ \hline
        EXAMS~\citep{hardalov2020examsmultisubjecthighschool} &  &  &  & \checkmark &  &  & Normalized accuracy \\ \hline
        FQuAD~\citep{dhoffschmidt2020fquadfrenchquestionanswering} &  & \checkmark &  &  &  &  & F1 score \\ \hline
        HellaSwag~\citep{zellers2019hellaswagmachinereallyfinish} &  &  &  &  &  & \checkmark & Normalized accuracy \\ \hline
        M3Exam~\citep{zhang2023m3exammultilingualmultimodalmultilevel} & \checkmark &  &  &  &  &  & Normalized accuracy \\ \hline
        Meta MMLU~\citep{llama3} &  & \checkmark & \checkmark &  &  &  & Normalized accuracy \\ \hline
        Mintaka~\citep{sen2022mintakacomplexnaturalmultilingual} &  & \checkmark & \checkmark &  &  &  & F1 score \\ \hline
        MLMM ARC~\citep{lai-etal-2023-okapi} &  & \checkmark & \checkmark & & \checkmark &  & Normalized accuracy \\ \hline
        MLMM HellaSwag~\citep{lai-etal-2023-okapi} & \checkmark & \checkmark & \checkmark & \checkmark & \checkmark &  & Normalized accuracy \\ \hline
        MLQA~\citep{lewis2020mlqaevaluatingcrosslingualextractive} &  &  &  & \checkmark &  &  & F1 score \\ \hline
        MMLU~\citep{hendrycks2020measuring} &  &  &  &  &  & \checkmark & Normalized accuracy \\ \hline
        OCNLI~\citep{hu-etal-2020-ocnli} & \checkmark &  &  &  &  &  & Normalized accuracy \\ \hline
        OpenBookQA~\citep{mihaylov2018suitarmorconductelectricity} &  &  &  &  &  & \checkmark & Normalized accuracy \\ \hline
        PIQA~\citep{bisk2019piqareasoningphysicalcommonsense} &  &  &  &  &  & \checkmark & Normalized accuracy \\ \hline
        SOQAL~\citep{mozannar2019neuralarabicquestionanswering} &  &  &  & \checkmark &  &  & Normalized accuracy \\ \hline
        TriviaQA~\citep{joshi-etal-2017-triviaqa} &  &  &  &  &  & \checkmark & Quasi-exact match \\ \hline
        TyDi QA~\citep{clark2020tydiqabenchmarkinformationseeking} &  &  &  & \checkmark &  &  & F1 score \\ \hline
        WinoGrande~\citep{sakaguchi2019winogrande} &  &  &  &  &  & \checkmark & Normalized accuracy \\ \hline
        X-CODAH~\citep{lin-etal-2021-common} & \checkmark & \checkmark & \checkmark & \checkmark &  &  & Normalized accuracy \\ \hline
        XCOPA~\citep{ponti-etal-2020-xcopa} & \checkmark &  &  &  &  &  & Normalized accuracy \\ \hline
        X-CSQA~\citep{lin-etal-2021-common} & \checkmark & \checkmark & \checkmark & \checkmark &  &  & Normalized accuracy \\ \hline
        XNLI 2.0~\citep{upadhyay2023xnli20improvingxnli} &  & \checkmark & \checkmark & \checkmark &  &  & Normalized accuracy \\ \hline
        XStoryCloze~\citep{DBLP:journals/corr/abs-2112-10668} & \checkmark &  &  & \checkmark &  &  & Normalized accuracy \\ \hline
        XWINO~\citep{tikhonov2021itsheadsusingattention} & \checkmark &  &  &  &  &  & Normalized accuracy \\ \hline
    \end{tabular}
}
\end{center}
\end{table}

\section{Average Rank Computation}\label{app:average_rank_comp}
Analogous to the method in FineTasks~\citep{kydlicek2024finetasksmultilingualtasks}, we compute the average rank for our ablations as follows:
\begin{enumerate}
    \item We train a model for each parameter configuration we want to ablate on.
    \item We evaluate each model on all the selected benchmarks.
    \item We compute the rank of each model (individual experiment) with regard to each benchmark and language.
    \item We compute the average rank for each model across all benchmarks and languages.
\end{enumerate}

\clearpage
\section{FineWeb documents in different scoring approaches}\label{app:examples}
To illustrate the types of documents each classifier scores highly or poorly, we present the highest- and lowest-scoring FineWeb examples for each of our classifier approaches (\emph{FT MKC$^+$}, \emph{MLP MKC$^+$}, \emph{CS MKC$^+$}). These examples were selected from the randomly chosen FineWeb test dataset (10K samples) used to validate the training of our model-based classifiers.

\subsection{FastText Classifier (FT)}
\begin{tcolorbox}[colback=color_blind_green!10!white,colframe=color_blind_green!90!black,title=Highest score:]
hi. i couldn't solve my problem because it has two conditional logical propositions. the problem is:can anyone help me about this, thanks =)we're expected to know that: . is equivalent tofind a logically equivalent proposition for:by first writing its contrapositive, and then applying demorgan's lawand the equality forthey were trying to be helpful by outlining the steps we should follow,. . but i think they made it more confusing.i don't see the purpose of using the contrapositive here.. . i wouldn't have done it that way.besides, the statement is a tautology . . .which gives us: .and this is a tautology: "a thing implies itself" ... which is always true.i don't know of any "logically equivalent proposition" we can write . . .
\end{tcolorbox}

\begin{tcolorbox}[colback=color_blind_red!10!white,colframe=color_blind_red!95!black,title=Lowest score:]
\UseRawInputEncoding
\begin{lstlisting}[language=, basicstyle=\normalfont, breaklines=true, columns=fullflexible, numbers=none, frame=none, xleftmargin=0pt, xrightmargin=0pt, aboveskip=0pt, belowskip=0pt]
|starts||23 sep 2016 (fri) (one day only)|want to travel soon but don’t wish to fork out a fortune for flights? check out today’s promotion from jetstar featuring promo fares fr $35 all-in valid for travel period commencing 12 october 2016don’t miss out! all-in frenzy fares to hong kong, penang and more from $35.sale ends 23 sep, 11pm!|travelling||price||travel period||find flight||penang||$35^|| [...]
\end{lstlisting}
\end{tcolorbox}

\subsection{Multi-Layer Perceptron (MLP)}
\begin{tcolorbox}[colback=color_blind_green!10!white,colframe=color_blind_green!90!black,title=Highest score:]
Naqhadeh County is a county in West Azerbaijan Province in Iran. The capital of the county is Naqadeh. At the 2006 census, the county's population was 117,831, in 27,937 families. The county is subdivided into two districts: the Central District and Mohammadyar District. The county has two cities: Naqadeh and Mohammadyar.
\end{tcolorbox}

\begin{tcolorbox}[colback=color_blind_red!10!white,colframe=color_blind_red!95!black,title=Lowest score:]
Custom Wedding Gifts

Personalized photo frames, albums \& keepsakes. Heirloom quality!

Custom Engraved Journals

Handmade in Florence Italy. Dozens of sizes and paper styles!

Awesome Leather Journals

Personalized, Customizable, Artisan made in Santa Fe, NM.

Ink Rendering from Photos

100\% Hand painted with unique style by pro artists. From \$49.
\end{tcolorbox}

\subsection{Cosine Similarity (CS)}
\begin{tcolorbox}[colback=color_blind_green!10!white,colframe=color_blind_green!90!black,title=Highest score:]
When you are renting a 5, 10, 15, 20, 30 or 40 yard dumpster, you want a company you can trust with prices that make you smile. Give us a call today and see the difference we can make in your next construction or clean out project.

Simply give us a call and we will help you figure out your dumpster rental needs.

Our dumpsters usually go out same-day or next-day depending on when you call.

We provide top-notch service, while going easy on your bottom line. What more could you ask for?

Our trained operators are here to give you a fast and hassle-free experience from start to finish.[...]
\end{tcolorbox}

\begin{tcolorbox}[colback=color_blind_red!10!white,colframe=color_blind_red!95!black,title=Lowest score:]
Cooperative flat 206/J

- Cooperative flat 201/J - Sold

2(1)+kitchenette, 50,1 m2Cooperative flat 202/J - Sold

2(1)+kitchenette, 44,9 m2Cooperative flat 203/J - Sold

2(1)+kitchenette, 50,6 m2Cooperative flat 204/J - Sold

1+kitchenette, 27,1 m2Cooperative flat 205/J - Sold

2(1)+kitchenette, 50,1 m2Cooperative flat 206/J - On sale

3+kitchenette 86,7 m2[...]
\end{tcolorbox}

\section{Example of a contaminated document}\label{app:decontmination}
We present an example of a FineWeb document that was removed during our decontamination pipeline.
\begin{tcolorbox}[title=MMLU contaminated document (matched 13-gram in bold):]
Here is our diagram of the Preamble to the Constitution of the United States. It is based on our understanding of the use of "in order to" as a subordinating conjunction that introduces a series of infinitival clauses (without subjects) that, in turn, modify the compound verbs "do ordain" and "establish."

See A Grammar of Contemporary English by Randolph Quirk, Sidney Greenbaum, Geoffrey Leech, and Jan Svartvik. Longman Group: London. 1978. p. 753.

We the People of the United States, in Order to form a more perfect Union, establish Justice, insure domestic Tranquility, \textbf{provide for the common defence, promote the general Welfare, and secure the Blessings} of Liberty to ourselves and our Posterity, do ordain and establish this Constitution for the United States of America.

If you have alternative rendering for this sentence, we would be happy to hear of it. Use the e-mail icon to the left.
\end{tcolorbox}

\section{License Information}\label{app:licenses}

\subsection{Dataset Licenses}

We use the following pretraining datasets:
\begin{itemize}
    \item FineWeb-2 (ODC-By license)
    \item FineWeb (ODC-By license)
    \item FineWeb-Edu (ODC-By license)
    \item DCLM (CC-BY 4.0)
\end{itemize}

We use the following classifier training datasets:
\begin{itemize}
    \item Aya Collection (Apache 2.0 license)
    \item Aya Dataset (Apache 2.0 license)
    \item Translated multilingual MMLU~\citep{openai2024mmmlu} (MIT license)
    \item OpenAssistant-2 (Apache 2.0 license)
    \item Include-Base-44 (Apache 2.0 license)
\end{itemize}

\subsection{Code Licenses}

We use the following open source code:

\begin{itemize}
    \item Nanotron (Apache 2.0 license)
    \item Datatrove (Apache 2.0 license)
    \item Lighteval (MIT license)
    \item FastText (MIT license)
\end{itemize}

\subsection{Model Licenses}

We use the following models:

\begin{itemize}
    \item Mistral v3 (Tekken) (Apache 2.0 license)
    \item XLM-RoBERTa (MIT license)
\end{itemize}

\end{document}